# PREDICTING LISTING PRICES IN DYNAMIC SHORT TERM RENTAL MARKETS USING MACHINE LEARNING MODELS

SAM CHAPMAN, SEIFEY MOHAMMAD AND KIMBERLY VILLEGAS

## TABLE OF CONTENTS

# PREDICTING LISTING PRICES IN DYNAMIC SHORT TERM RENTAL MARKETS USING MACHINE LEARNING MODELS

**Sam Chapman, Seifey Mohammad and Kimberly Villegas**
Department of Computer Science
Rice University
{sjc8,sum1,kv22}@rice.edu

**Dr. Yu, Cheng**
Rice University Mentor
yc180@rice.edu

## ABSTRACT

Our research group wanted to take on the difficult task of predicting prices in a dynamic market. And short term rentals such as Airbnb listings seemed to be the perfect proving ground to do such a thing. Airbnb has revolutionized the travel industry by providing a platform for homeowners to rent out their properties to travelers. The pricing of Airbnb rentals is prone to high fluctuations, with prices changing frequently based on demand, seasonality, and other factors. Accurate prediction of Airbnb rental prices is crucial for hosts to optimize their revenue and for travelers to make informed booking decisions. In this project, we aim to predict the prices of Airbnb rentals using a machine learning modeling approach.

Our project expands on earlier research in the area of analyzing Airbnb rental prices by taking a methodical machine learning approach as well as incorporating sentiment analysis into our feature engineering. We intend to gain a deeper understanding on periodic changes of Airbnb rental prices. The primary objective of this study is to construct an accurate machine learning model for predicting Airbnb rental prices specifically in Austin, Texas. Our project's secondary objective is to identify the key factors that drive Airbnb rental prices and to investigate how these factors vary across different locations and property types.

1. INTRODUCTION

The emergence of Airbnb has transformed the travel industry, allowing homeowners to rent out their properties to travelers seeking unique and affordable accommodations. The pricing of Airbnb rentals is highly dynamic, influenced by factors such as demand, seasonality, and location-specific attributes. Accurate prediction of rental prices is essential for hosts to optimize their revenue and for guests to make well-informed booking decisions. This project presents a comprehensive study on predicting Airbnb rental prices using machine learning models, with a specific focus on the vibrant city of Austin, Texas.

1.1 BACKGROUND

In recent years, the sharing economy has seen significant growth, thanks to platforms like Airbnb that have revolutionized the accommodation industry. The sharing economy involves individuals sharing resources, goods, or services through online platforms or communities, which optimizes the use of underutilized assets.

With Airbnb playing a crucial role in transforming how people search and book accommodations, one of the primary challenges that arise is the fluctuating nature of Airbnb prices. Unlike traditional accommodations, Airbnb prices can vary based on factors such as location, demand, seasonal fluctuations, and unique listing characteristics. This unpredictability poses a distinct challenge for hosts and guests who strive to find the right balance between affordability and value.

To address this challenge, the integration of data science has become increasingly important. Researchers can leverage the vast amount of historical data available on Airbnb to gain valuable insights into pricing trends, customer preferences, and market dynamics. These insights serve as the building blocks for creating accurate models capable of predicting Airbnb listing prices.

Accurate prediction of Airbnb listing prices holds significant implications and contributions for both hosts and guests within the sharing economy. For hosts, precise price predictions empower them to optimize their listings by setting competitive prices that reflect market dynamics and property attributes. This, in turn, maximizes their revenue potential and enhances their overall business performance. It also provides hosts with insights into the factors driving price variations, allowing them to make informed decisions regarding property upgrades, amenities, and many other listing attributes.

On the other hand, accurate price predictions benefit guests within the sharing economy by providing transparency and enabling them to make informed decisions based on their budget, preferences, and desired amenities. Within the sharing economy, where resources are shared among individuals, accurate price predictions allow guests to select the most suitable accommodation options and plan their trips effectively. This transparency enhances their overall satisfaction and experience of participating in the sharing economy through the Airbnb platform.

We wanted to apply this concept of price prediction regarding Airbnb rentals to the housing market of Austin, Texas. Austin is an area of the United States that has seen significant growth in the last decade, especially over the COVID-19 pandemic; it experienced a large influx of people and businesses moving there. On top of all that, Austin is the home to large exclusive events such as Austin City Limits (ACL) and South by Southwest (SXSW). All of which have led to a very dynamic and fluctuating short term rental market. We hope to study this market and derive accurate and helpful insights with this paper.

1.2 OBJECTIVES

The primary objectives of this project are:

- Predict Airbnb listing prices in Austin, Texas, using various models: Develop accurate prediction models specifically tailored to the Airbnb market in Austin, Texas. Employ models such as linear regression, random forest, and decision trees to generate predictions that capture the trends and fluctuations in listing prices. These models will offer valuable insights to hosts and potential guests, which can enable them to gain a deeper understanding of the expected price trends in Austin's dynamic short-term rental market.

- Identify the key factors that significantly impact listing prices in the city: Examine, recognize, and evaluate the significant elements that can potentially impact the prices of Airbnb listings in Austin, Texas. This knowledge allows hosts to optimize their pricing strategies based on specific property features, geographical positioning, seasonal trends, reviews and local attractions. Likewise, guests can make informed choices, taking into account the factors that drive prices and tailor their accommodation to align with their preferences and budget.

2. RELATED WORKS

Accurately predicting prices is extremely important in the short-term rental market, especially for platforms like Airbnb. Previous studies have delved into creating models that can forecast the prices of Airbnb listings. These studies have used different machine learning algorithms, such as linear regression and random forest, to gain insights into pricing patterns and determine the main factors that can impact listing prices. In this section, we will take a closer look at the existing literature on price prediction in the accommodation industry, focusing on regression-based approaches.

2.1 LINEAR REGRESSION

Linear regression is an algorithm that provides a linear relationship between a dependent variable (price) and one or more independent variables (features). It estimates the coefficients to minimize the sum of squared differences between the predicted and actual prices. The general equation for a simple linear regression with one independent variable is:

$$Y = β_0 + β_1 X + ϵ \qquad (1)$$

Where, Y is the dependent variable (in this case, Airbnb price). X is the independent variable (a factor influencing price, such as, room type, number of bedrooms, etc.). $β_0$ is the y-intercept (the value of Y when X is 0). $β_1$ is the slope of the regression line which represents the change in Y for a unit change in X. $ϵ$ is the error term which captures the deviation of the actual values from the predicted values.

$$Y = β_0 + β_1 X_1 + β_2 X_2 + … + β_n X_n + ϵ \qquad (2)$$

The equation can be extended for multiple independent variables (denoted as $X_1, X_2, …, X_n$) by adding additional terms with their corresponding coefficients ($β_2 X_2, β_3 X_3, …, β_n X_n$) as can be seen on equation 2.

Advantages of Linear Regression:

- Linear regression performs well for linearly separable data, assuming there is a linear relationship between dependent and independent variables.
- It offers ease of implementation, interpretation, and computational efficiency compared to more complex algorithms.
- Linear regression can serve as a baseline model to compare against more complex models.

Disadvantages of Linear Regression:

- Linear regression is susceptible to the presence of noise and overfitting, which may compromise the model's predictive accuracy.
- It is sensitive to outliers, as these extreme values can exert disproportionate influence on the estimated coefficients and overall model performance.
- It is prone to multicollinearity, a condition in which highly correlated independent variables can introduce instability in the coefficient estimates.

Previously, Yu et al. (2018) conducted a study titled "Real Estate Price Prediction with Regression and Classification." In their work, they utilized linear regression as their based line model to forecast a home's potential sale price. They used a dataset with 288 features and 1,000 training samples. The linear regression model was trained using all features, and the predicted sale prices were compared to the actual sale prices in the test dataset. The performance of the model was measured using Root Mean Square Error (RMSE). RMSE is one of the most commonly used measures for evaluating the quality of predictions. It shows how far predictions fall from measured true values using Euclidean distance. Lower RMSE values indicate better predictive accuracy, as they represent smaller differences between predicted and actual values. The baseline linear regression model generated an RMSE of 0.5501.

To address overfitting, the authors incorporated regularization parameters in the linear regression models. They used Least Absolute Shrinkage and Selection Operator (Lasso) regularization and performed 5-fold cross-validation. Lasso adds a penalty term to the loss function, which encourages the model to shrink the coefficients of less important features towards zero. This promotes feature selection and prevents overfitting. Cross-validation is a technique used to assess the performance of a machine learning model. It involves dividing the available data into multiple subsets or folds. The model is then trained on a portion of the data (training set) and evaluated on the remaining portion (validation set). This process is repeated multiple times, with each time with a different split of the data. The Lasso regularized linear regression model resulted in an improved RMSE of 0.5418. The model also automatically selected 110 variables and eliminated the other 178 variables to fit the model.

The authors then applied Ridge regularization with cross-validation in their linear regression model. Similar to the lasso regression, Ridge regression puts a similar constraint on the coefficients by introducing a penalty factor on the square of the coefficients. It reduces the magnitude of coefficients without forcing them to zero, thereby addressing overfitting. The Ridge regularized linear regression model produced an RMSE of 0.5448, further demonstrating the benefits of regularization in reducing overfitting.

The authors successfully demonstrated the effectiveness of incorporating regularization techniques, such as Lasso and Ridge regularization, to enhance prediction performance and address overfitting in the domain of house price predictions. These findings confirm that regularization techniques play a crucial role in improving prediction accuracy and mitigating overfitting in price prediction models.

Similarly, in the context of price prediction for Airbnb listings, previous works have explored the use of linear regression to analyze the relationship between various factors and listing prices. One such study conducted by Yang et al. (2016) set out to investigate how market accessibility impacts hotel prices in the Caribbean region. Recognizing that market accessibility alone may not be the sole determinant of prices, the researchers also considered additional contributing factors, namely user ratings and hotel classes, which were hypothesized to exert a moderating effect on the relationship between market accessibility and prices.

The study revealed that user ratings played a significant role in influencing hotel prices. Positive online ratings were found to have a positive impact on room rates, indicating that guests were willing to pay more for accommodations with higher user ratings. The study showed that even a one-point increase in online reviewer rating was associated with a considerable hotel price premium of 53.0%. These findings suggest that user ratings serve as a major factor in the hotel industry. Guests are willing to pay a higher price for accommodations that have received positive feedback from previous guests. Which indicates the importance of reputation and customer satisfaction in determining hotel prices.

Elastic Net, is another regularization technique that blends both Lasso and Ridge regularization. It strikes a balance between their strengths, providing a versatile regularization approach. In the study conducted by Mat et al. (2023) investigated the factors influencing the consumer price of milk in Turkey using Elastic Net. This study aimed to identify the structure of basic and economic indicators affecting milk prices. The authors employed time series data analysis from 2010 to 2020, revealing that the Elastic Net model explained 97.8% of the variations in milk prices. The findings highlighted the relationship between consumer milk prices and variables such as producer price of milk, feed prices, exchange rates, and economic indices.

The incorporation of Elastic Net regularization into price prediction models for Airbnb listings holds significant promise. Just as Elastic Net effectively captured the complex interactions among various factors impacting milk prices, it could similarly capture intricate relationships within the Airbnb market.

Linear regression, when augmented with techniques like Lasso, Ridge, and Elastic Net, allows for enhanced predictive accuracy while mitigating the risks of overfitting. The incorporation of regularization techniques alongside the consideration of user ratings shows potential in enhancing the accuracy of price prediction models for Airbnb listings and can provide valuable insights into the factors influencing pricing decisions. Taking into account online reviews and user feedback, property hosts can gain a deeper understanding of how their property's reputation influences pricing strategies. Similarly, prospective guests can rely on user ratings as a reliable gauge of quality when making informed decisions about accommodations.

2.2   RANDOM FOREST

Random Forest Regression is a machine learning algorithm that combines the principles of ensemble learning and decision trees to perform regression tasks. Decision trees are a machine learning algorithm that partitions data into hierarchical branches based on feature values to make predictions or decisions. In decision tree regression, each internal node represents a feature test, and leaf nodes hold predicted values, enabling the tree to estimate continuous target variables by traversing the tree based on feature conditions. So the Random Forest algorithm is an extension of the Decision Tree. The main idea behind Random Forest Regression is to create multiple decision trees and average their predictions to obtain a more accurate and robust regression model.

The algorithm works as follows: The training data is divided into random subsets (bootstrap samples) through a process called bagging. Each subset is used to train an individual decision tree. Next is tree construction, for each bootstrap sample, a decision tree is built by recursively partitioning the data based on various features and their corresponding split points. The tree is grown until a specified stopping criterion is reached, such as a maximum depth or minimum number of samples required to split a node.

In the past, Adetunjia et al. (2022) used the Random Forest Regression in a study titled, "House Price Prediction using Random Forest Machine Learning Technique." This paper examines the application of the Random Forest machine learning approach in regards to the Boston housing market, the study assesses its efficacy in predicting home prices. To evaluate the performance of the proposed prediction model, the Boston housing dataset from the UCI Machine Learning Repository, comprising 506 entries and 14 features, was utilized. Features included: crime rate per capita, property tax rates, median value of owner-occupied homes, and so forth.[1] The proposed model was created using the random forest

---
[1] Please refer to *Adetunjia, et al. "House Price Prediction using Random Forest Machine Learning Technique"* for a full list of features in section 2.1

algorithm, utilizing the RandomForestClassifier from the Python Scikit-learn library. Random Forest is a popular supervised machine learning technique for classification and regression tasks. It employs ensemble learning, combining multiple decision trees to improve the accuracy of predictions. By averaging the outcomes of these individual trees applied to different subsets of the dataset, Random Forest enhances the overall predictive accuracy. Instead of relying on a single decision tree, the final prediction is based on the majority of votes from the ensemble of trees, resulting in a more robust and reliable model.

Once their model was trained using the training dataset, the subsequent stage of the study involves testing the model's predictive capability. This was accomplished by excluding the actual prices from the dataset and using the model to simulate predictions for house prices. The predicted prices and actual prices were then compared and the differences were calculated. The results demonstrated that, while exact prices were not always predicted, the variances between the predicted values and the actual values fell within the range of ±5. Their model also produced an RMSE of approximately 2.5 and an R-squared of 0.9.

Furthermore, the research done by Hu et al. (2023) in the paper "Prediction and Analysis of Rental Price using Random Forest Machine Learning Technique" looks at the rental prices within Shanghai and Wuhan and attempts to study the price trend of the housing rental market and attempts to build a model that is robust in predicting the rental prices. In terms of collecting their data they collected data from a Chinese real estate company LIANJIA for 2022 rental data. They had 2546 rows of data for Shanghai and 2377 rows of data for Wuhan, and they performed a standard 80-20 train test split for their data. The data had features such as the number of bedrooms, living rooms, bathrooms, total area, and other intrinsic characteristics of the property.

Next, once the model was constructed and trained they ran it against the test data and saw that the Random Forest regression model for the Shanghai data had an MAE (Mean Absolute Error) of about 922.16 and an r-squared of 0.83. For the Wuhan data it had an MAE of about 234.71 and an r-squared of 0.91. This research reveals that the efficacy of the random forest model in predicting rental prices in Shanghai is unsatisfactory due to the high MAE. Despite the selected variables exhibiting a strong explanatory impact on rental prices in the city, the average disparity between the projected and actual values remains substantial. The researchers posit that the variables chosen for this study may not provide a comprehensive elucidation of the rental prices in Shanghai. In the case of prominent cities like Shanghai, rental prices could potentially be influenced by a broader array of factors and confounding variables. So the researchers propose that a future update to their paper would be to incorporate far more features and data points. In the context of Wuhan, the random forest model demonstrates a notably superior predictive capability for rental house prices. The alignment between the projected prices and actual prices is strong, underscoring the effectiveness of the variables chosen in this study to aptly elucidate rental house prices in Wuhan.

Advantages of Random Forest:

- Robustness to Overfitting: Random Forest regressors are less prone to overfitting compared to individual decision trees. By aggregating the predictions from multiple trees, the ensemble model reduces the impact of outliers and noise in the dataset, resulting in more reliable predictions.
- Handling Non-linear Relationships: Random Forest regressors can capture complex non-linear relationships between the input features and the target variable. They are capable of automatically learning and incorporating interactions and non-linearities present in the data, making them suitable for modeling housing price prediction where multiple factors influence the outcome.

Disadvantages of Random Forest:

- Computational Complexity: Random Forest regressors can be computationally expensive, especially when dealing with large datasets or a large number of decision trees in the ensemble. Training and predicting with a Random Forest regressor can require substantial computational resources, which may limit their practicality in certain scenarios.
- Sensitivity to Noisy Data: Random Forest regressors are sensitive to noisy or irrelevant features in the dataset. If the input data contains noisy or misleading variables, the Random Forest model may still incorporate them in the learning process, leading to suboptimal predictions. Proper feature selection and data preprocessing techniques are crucial to mitigate this issue.

## 2.3 EXTREME GRADIENT BOOSTING (XGBOOST)

Extreme Gradient Boosting (XGBoost) is a powerful machine learning algorithm that combines the principles of gradient boosting with enhanced efficiency and flexibility. Gradient boosting is a machine learning technique where weak models (typically decision trees) are trained sequentially to correct errors made by previous models. XGBoost, an optimized implementation of gradient boosting, improves performance through parallel tree construction, regularization, and handling missing values, resulting in a powerful and efficient predictive model. It is widely used for regression and classification tasks due to its high performance and interpretability.

At its core, XGBoost aims to minimize a loss function by sequentially adding decision trees that correct the errors made by the previous trees. The algorithm's key steps are as follows: first, initialize the model with a constant value, usually the mean of the target variable. Second, for each iteration (or boosting round), calculate the gradients with respect to the current predictions. Third, build a weak decision tree to fit the negative gradients of the loss function (residuals). This tree represents the new contribution to the model. Fourth, Update the predictions by adding the predictions of the newly created tree, scaled by a learning rate. Repeat steps 2 through 4 until a predefined number of trees or a convergence criterion is met. Finally, the predictions of all the trees are combined to obtain the final prediction.

In 2020, Rolli et al. used an XGBoost regression in a study titled, "Zillow Home Value Prediction (Zestimate) By Using XGBoost." The objective of this paper is to construct an XGBoost model for housing price prediction utilizing various features such as the number of bedrooms, bathrooms, kitchen size, and square footage. The dataset, sourced from Kaggle.com, encompasses property listings from three California counties that are near the Los Angeles metropolitan area.

Their model was trained using the inherent characteristics of the home as features, so a mix of qualitative and quantitative variables, and the log-error of the home was used as the target variable.

Where $log - error = log(Zestimate) - log(sale\ price)$, and ultimately using negative mean absolute error as their performance indicator they saw -0.05949 for their XGBoost model. Ultimately seeing this as a very accurate model.

Advantages of XGBoost Regression Model:

- High Predictive Performance: XGBoost is known for its exceptional predictive accuracy. It handles complex relationships between features and target variables effectively, making it suitable for tasks where accurate predictions are crucial.
- Robust to Overfitting: XGBoost incorporates regularization techniques like shrinkage and pruning, which help prevent overfitting. It can handle noisy data and complex datasets, providing reliable predictions even with limited training examples.

Disadvantages of XGBoost Regression Model:

- Computational Complexity: XGBoost can be computationally intensive, especially when dealing with large datasets or complex models. Training an XGBoost model may require more time and computational resources compared to simpler algorithms.
- Model Complexity and Interpretability: As an ensemble of decision trees, the resulting XGBoost model can be complex and challenging to interpret. The individual decision trees can be visualized, but understanding the overall logic and feature importance can be more difficult compared to simpler models like linear regression.

## 3. DATA DESCRIPTION

The data for this algorithm was acquired from multiple sources and is merged together on one shared column. The datasets are sourced from three different platforms - insideairbnb.com, GitHub and Kaggle. Insideairbnb.com was the original source of the data until shortcomings occurred and additional sources were required to reach the goal of the project. GitHub and Kaggle were our next sources for data, but it was sourced originally via insideairbnb.com and saved on their respective platform for use in previous projects related to Airbnb. Insideairbnb.com has a paywall for access to historical data prior to the year 2023, so the access to their previous data limited its use in regards to calendar pricing information.

In its raw format, the datasets are split into three data frames - listings, calendar, and reviews. The listings data frame consists of descriptive information of a given Airbnb listing. The calendar data frame contains the price of a listing on specific calendar dates of the year. The reviews data frame consists of all the reviews posted on all the Airbnb listings in and around Austin, Texas since Airbnbs began in the area. The datasets are all connected via a unique 'listing_id' that is provided by Airbnb for a specific listing on their platform. Additional items verify the identity of the listing, but 'listing_id' is the unique identifier that connects across all datasets used in the project.

Following data preprocessing of the three main groupings of datasets; listings, calendar, and reviews; the final dataset used for the algorithm consists of 18,105,646 rows and 91 columns.

## 3.1 LISTINGS

Insideairbnb.com sources the listings data frame, from Inside Airbnb[2], which includes 14,368 rows and 74 columns in its raw format. This data frame contains all Airbnb listings within the Austin area and includes descriptive information such as: amenities, location, number of beds, property type, and various review scores. The data frame was scraped from Airbnb's website over the course of two dates - March 16th and 17th, 2023. This data is informational and static as it does not include time-based information.

The listings dataset is appropriate for study and analysis due to its large volume of information. For a given Airbnb listing, information like the validation of certain amenities, the location of the site, and the property type are crucial pieces of information for a pricing model among other things. Utilizing the more descriptive and detailed information of a specific listing is integral to the success of a pricing model when it comes to Airbnb properties.

The listings dataset used in this project has certain limitations that should be acknowledged. One key limitation is that the price information provided for each listing represents a specific point in time when the data was scraped from insideairbnb.com. This static price does not adequately capture the dynamic nature of listing prices, which can fluctuate significantly throughout the year, even on a daily or weekly basis. To address this limitation, we incorporated the calendar dataset, which provides more granular pricing information and takes into account temporal variations. Another challenge encountered in the dataset was the presence of missing values (NaN) in various columns, such as 'bed' and 'property type'. To resolve this, a manual approach was used to retrieve accurate information by referencing the provided URL for each listing. For cases where the number of missing values was substantial, alternative methods were employed. For instance, the 'review_scores_location' feature, which contained several NaN values, was processed using a nearest neighbor function to impute missing values. It is important to note that while this approach has been effective in resolving missing values, there may be instances where the assumption of proximity to neighbors may not hold true. As a result, further improvements will be made to enhance the accuracy of this imputation method.

---

[2] Inside Airbnb website: http://insideairbnb.com/get-the-data

Table 1: A subset of the listings dataset post-processing.

| ID | Host ID | Host is Superhost | Host Listing Count | Host Total Listings | Neighborhood Cleansed | Latitude | Longitude |
|---|---|---|---|---|---|---|---|
| 5456 | 8028 | 1 | 2 | 4 | 78702 | 30.260570 | -97.734410 |
| 5769 | 8186 | 1 | 1 | 4 | 78729 | 30.456970 | -97.784220 |
| 6413 | 13879 | 1 | 1 | 1 | 78704 | 30.248850 | -97.735870 |
| 6448 | 14156 | 0 | 1 | 2 | 78704 | 30.260340 | -97.764870 |

### 3.2 CALENDAR

The majority of the calendar dataset is sourced from GitHub and Kaggle, which was originally sourced from insideairbnb.com. The data is sourced over the course of several years - 2019 through 2024. As noted above, Inside Airbnb has restrictions over historical data, so the other two platforms were used for the previous calendar information as it was used for Airbnb-related projects on their platform and is available for public use. The current year of data is publicly available on the insideairbnb.com website and so was extracted for use in the full calendar dataset.

Following the merge of the datasets, the data frame consists of 23,697,989 rows and 7 columns. The columns of the data frame represent information such as: listing_id, date, available, price, minimum_nights, and others as seen in Table 2 below. This full dataset ranged from 09/19/2019 through 03/29/2024, but there were some NaNs present and duplicates that were dealt with. There is missing over one section of time from 04/22/2021 to 06/07/2022. For the gap in the data, a separate data frame was made of the average price for each individual listing by each day of the week. These prices were then filled in for the gap in data. At the conclusion of the merge of the several years worth of data and removing duplicates, the data frame consists of 33,290,877 rows and 4 columns.

This dataset is appropriate for the project and overall goal due to the ability to represent how the price of a listing changes over time down to the day of the week. Through exploratory analysis seen below, it was found that the price fluctuates depending on season of the year, month, and day of the week, so the group decided it was necessary to incorporate the pricing over a period of time to more accurately predict a price of an Airbnb listing.

Limitations and drawbacks for the calendar sourced data fall into two mentionings. The data was sourced via secondary sources rather than the original source of insideairbnb.com. Unfortunately, the group was unsuccessful in acquiring the historical data directly through their platform; that being said, the data is formatted exactly the same as it is for the current 2023 year's data that is publicly available on the website. The second drawback is for the missing section of data mentioned above. That missing section was dealt with in a manner the group deemed appropriate to reach the goal for the project.

Table 2: Subset of calendar dataset post-processing.

| Listing ID | Date | Price | Month |
|---|---|---|---|
| 5456 | 2022-06-09 | 95 | 6 |
| 5456 | 2022-06-10 | 150 | 6 |
| 5456 | 2022-06-11 | 150 | 6 |
| 5456 | 2022-06-12 | 95 | 6 |

3.3  REVIEWS

The reviews dataset was acquired through insideairbnb.com. In its raw format, it consists of 944,636 rows and 4 columns. The columns represent the listing id, the id of the specific review, the date the review was posted, and the full comment from the reviewer. Reviews are made after a guest has stayed at the Airbnb listing and is posted on the listing's page.This information was scraped by insideairbnb.com and made into a .csv file, which is updated periodically to add more recent reviews. The data range dates back from March 17, 2009 to March 16, 2023. Since the comments cannot be quantified in their raw state, sentiment analysis is performed to result with a numerical value for further analysis in the model.

The reviews are valuable to the data science problem as they provide important feedback to how a guest feels about a given listing. It allows for additional insight to the listing beyond the standard ratings out of five stars for specific aspects of the listing. The group feels that with the increasing prevalence of reviews being posted on the listings around Austin, the reviews are necessary to differentiate a listing from another that is similar to it. Additionally, as a guest reviews are important feedback and may boost favor of one listing over another if all the reviews are positive.

Limitations and drawbacks for the reviews data come down to a few issues worth noting. Some of the reviews are not written in english, so the sentiment analysis is unable to transcribe this into numerical value and thus they are dropped. Additionally, the reviews date back to dates prior to the data ranges for the calendar data used in the project. The reviews prior to the first calendar date may falsely represent a listing as the owner may have taken steps to improve the listing since those reviews were posted.

Table 3: Subset of reviews dataset in its raw form.

| Listing ID | ID | Date | Comments |
|---|---|---|---|
| 2265 | 963 | 2009-03-17 | I stayed here during SXSW and had a really pleasant stay.  The house is a very relaxing environment and Paddy is both a friendly and professional host. |
| 2265 | 200418 | 2011-03-16 | We had a great time in Austin staying at Paddy's house. The house is much larger than it looks on the photos, very clean and the moment you step in you feel like home. Wouldn't hesitate to stay there again next time I'm back in Austin. |
| 2265 | 1016390 | 2012-03-19 | I arrived late in the evening so did not meet Paddy, but her home was lovely and clean and welcoming. She checked in with my friend and I most days to make sure that we had no questions or needed anything. All in all, a very smooth, stress-free experience in Austin! |

4.  DATA SCIENCE PIPELINE

The data science pipeline section highlights the systematic and iterative process followed in this project, from data collection and preprocessing to model selection. The pipeline demonstrates the integration of domain knowledge, data analysis, and advanced modeling techniques to predict Airbnb rental prices accurately. By following this well-defined pipeline, insights gained from the project can be effectively applied to future price prediction tasks, benefiting both hosts and guests in the dynamic sharing economy of Airbnb rentals.

4.1  DATA WRANGLING

Data wrangling is an essential step in the data science pipeline process. In the context of predicting Airbnb prices, efficient data wrangling techniques are crucial for cleaning and organizing the datasets. This section outlines the steps taken to perform data wrangling on the different datasets, including merging different datasets, removing irrelevant data, and handling missing values and outliers.

### 4.1.1 PREPROCESSING LISTINGS

The listings dataset used in the model was obtained from one source - insideairbnb.com. Due to the singular source, there is no need to merge the dataset with others prior to preprocessing. Note, it is merged with the calendar and review data later on. Furthermore, the task of preprocessing the listing dataset comes down to three main steps - condensing, feature engineering, and removal.

The dataset in its raw format has 74 column features. While 74 features for the model is a desirable value, the features present on the raw import are not all important for the prediction model. Many of the features need to be removed and that was done manually by parsing through and dropping the columns that were not of interest in regards to the model. Columns that were fit for removal were columns such as; 'scrape_id,' 'listing_url,' and a total of thirty six columns were removed to condense the data frame down to thirty nine columns before feature engineering took place.

Data engineering was carefully incorporated into the preprocessing of the data frame. The latitude and longitude coordinates are included in the listings file, and the potential was seen in adding in locational features dependent on the certain locations of interest in and around Austin. Through research, a list of the top fifthteen most popular tourist attractions was gathered. The locations involved the state capitol, sight-seeing spots, museums, parks, and more. To involve the destinations, the longitude and latitude coordinates for locations of interest were extracted via GoogleMaps and added to a list. It is noted that two of the locations did not have a defined coordinate location, so they were removed from the list in case of bias. The list was then put through a distance calculation, with the package 'GeoPy,' for each of the listings, and the result appended a distance value in kilometers to each of the locations of interest. With this step, all the listings gained an additional thirteen features, each representing the distance to a popular location in Austin. Another significant increase in the features total came from the amenities column. In its raw format, the amenities column includes a list, however long, of each amenity declared by a host for a given listing. The list of amenities ranges dramatically per each listing, so to quantify this, a counting function was involved to see which amenities were most common across all the Airbnbs around Austin. Once this list was established, it was condensed down to the top thirty amenities most prevalent across all listings. Each listing was then compared with the top thirty list and binary columns were added to represent whether or not a given listing had each amenity.

Following feature engineering, there were NaN values present, and so removal and cleaning had to be done. A mixture of drops and feature manipulation had to be done. Feature manipulation in order to preserve the listings count as best as we could. The main source of the null values came across in relation to review scoring. Any guest has the option to leave a review score, out of five, after staying at the location, and the average of all of these scores are posted on a listing. The review ratings span across eight categories; overall rating, accuracy, cleanliness, checkin, communication, value, reviews per month, and location. Roughly 3,100 of the listings had missing review score values across these categories, so feature engineering and manipulation was carried out to resolve this issue. For all the review columns except location, two steps were carried out. A significant number of hosts have multiple properties, so we used the host id to find reviews of that particular host's other properties and replaced the null values with a mean of their other scores in the respective categories. This process resolved roughly 1,200 of the scores. For the rest, a median of all Austin listings was used to replace the missing scores. The location scores were resolved differently. Instead, a nearest neighbors function was carried out to calculate the average score of the ten nearest neighbors, or listings, and use a mean of the review location scores as a replacement value. This does raise an issue in regards to if the ten nearest listings are significantly far away from a particular listing. Future work is to involve a check verifying the distance is not outside of a sufficient range. Another significant source of null values came from the beds column, which states how many beds are inside a given listing. To resolve this, a team member manually went through the URLs of the listings missing this information and appended the accurate amount. Following these steps, any listings that still contained NaN values were dropped.

It was decided in order to maintain the rich diversity of Austin's listings, maintaining as many from the raw import was crucial to the model. These steps were necessary to prepare the listings data frame ahead of the merge with the calendar and reviews datasets. The resulting listings data frame has a shape of 14,345 rows and 86 columns.

### 4.1.2 PREPROCESSING CALENDAR

The calendar data for predicting Airbnb prices was obtained from multiple sources and merged. The individual datasets are concatenated vertically to create a unified dataset. To streamline the data, unnecessary columns are dropped and duplicates are removed. The removal of columns such as whether the listing is available, or the minimum number of nights necessary to book a specific listing, helps to simplify the dataset, by retaining only the essential information. All columns are then evaluated to remove any unnecessary characters such as "$" and ",". Missing values and outliers are then addressed using the following steps:

1. Create missing values dataframe: The dataframe is checked for missing values. Rows with missing values are isolated into a separate missing values dataframe.
2. Drop rows with missing values: Rows with missing values are removed from the original dataframe.
3. Handling outliers: Outliers are detected using the traditional Interquartile Range (IQR) method. Rows with prices outside the calculated thresholds are eliminated.

Outliers are data points that deviate significantly from the rest of the dataset. They can arise due to various reasons, such as measurement errors, data entry mistakes, or extreme values. Outliers can have a considerable impact on statistical measures such as the mean and standard deviation, leading to biased results.

IQR, is a statistical measure that represents the range between the first quartile (Q1) and the third quartile (Q3) in a dataset. In the IQR Method the quartiles divide a dataset into four equal parts, where the first quartile represents the 25th percentile and the third quartile represents the 75th percentile. IQR is often used as a basis for outlier detection. In this approach, outliers are identified as data points that fall below $Q1 - multiplier * IQR$ or above $Q3 + multiplier * IQR$. The multiplier acts as a constant to determine the threshold for discerning outliers. A larger multiplier value will result in a wider range and capture more data points as outliers, while a smaller multiplier value will result in a narrower range and fewer data points flagged as outliers. In this project, a multiplier of 0.5 is used. The choice of a smaller multiplier such as 0.5 indicates a more conservative approach to outlier detection. It will result in a narrower range of values being considered as outliers, thereby retaining more data points in the dataset. This approach is suitable when the aim is to preserve a larger portion of the data while still identifying potential outliers

The lower fence and upper fence correspond to the lower and upper thresholds, respectively. Detected outliers are then removed from the dataset.

Figure 1 visually demonstrates the impact of applying the IQR method. It presents the dataset before and after the removal of extreme outliers, highlighting the effect of the IQR-based outlier detection and removal process.

The calendar dataframe is missing data between April 22, 2021 to June 07, 2022. To address the missing dates, the following steps are performed:

1. Calculate average prices: Average prices per each unique listing is calculated for each day of the week on the original dataframe.
2. Create a new dataframe for missing time period: A new dataframe is created to include the missing time period: April 22, 2021 to June 07, 2022 with all the listings id's from the original dataframe.
3. Merge dataframes: The dataframe containing missing values (from previous section) and the new time period dataframe are concatenated.
4. Combine dataframes: The merged dataframe, from step 3, are inner-joined with the average listing prices, from step 1, to fill in the missing price values.

This expanded dataset provides a more extensive representation of Airbnb prices over time. Through merging datasets, removing irrelevant information, handling missing values and outliers, and filling missing dates, we obtained valuable insights and have prepared the data for subsequent analysis.

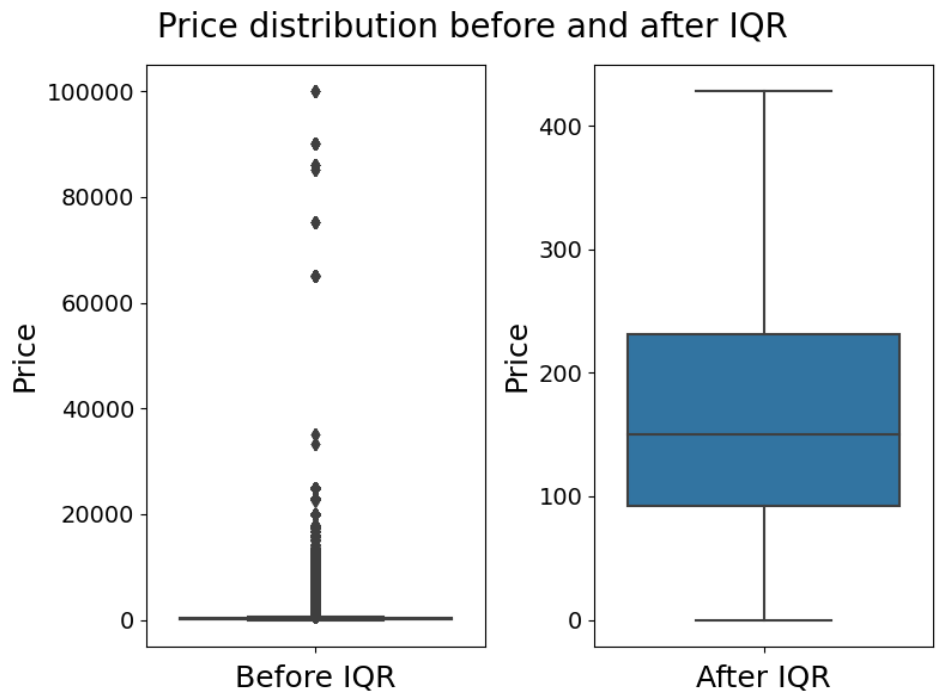

Figure 1: Price distribution depicting before and after IQR implementation

Before: The boxplot of prices shows a broader and more varied range, with values spanning from lower to higher price points. The whiskers extend to the minimum and maximum values, indicating the presence of extreme outliers. After: A significant reduction in the spread of prices is observed in the box plot, indicating that the implementation of the IQR technique has effectively filtered out extreme outliers from the dataset. The whiskers now represent the data within a reasonable range, while the box itself represents the interquartile range (IQR) containing the central 50% of the data. The median line indicates the central tendency of the distribution.

### 4.1.3 PREPROCESSING REVIEWS

In the case of the review's datasets, several data wrangling techniques are applied to ensure the data is clean, relevant, and ready for further analysis. The first step in data wrangling for the review data is to merge the different datasets. In this case, the review datasets from different years are combined to create a comprehensive dataset. The datasets are imported and read using the appropriate libraries and functions. Then, the datasets are merged vertically using the concatenation function to combine them into a single dataset. Relevant columns are selected for the merged dataset to ensure only necessary information is included.

To ensure the dataset contains relevant information, any irrelevant or unnecessary data needs to be removed. In the context of review data, one common step is to identify and handle missing values. Rows with missing values, particularly the review comment itself, are replaced with an empty string and separated into its own missing values dataset. Missing values in other columns besides the comment section are removed. All comments not in the English language are removed. This ensures that only complete and meaningful data remains for analysis.

Pre-processing is crucial to clean and normalize the review's text data. Several pre-processing steps are performed, such as removing URLs, HTML tags, extra spaces, as well as redundant punctuation (e.g., "!!" or "???"). Contractions are expanded to their full form to ensure accurate analysis. Emojis are replaced with their corresponding textual representations, and extra lines and tabs are removed. These pre-processing steps are essential in standardizing the text data and preparing it for feature engineering in sentiment analysis.

Sentiment analysis is a valuable technique for understanding the sentiment expressed in reviews. The Valence Aware Dictionary and Sentiment Reasoner (VADER) is a commonly used sentiment analysis tool that provides sentiment scores and categories for each review. VADER operates as a lexicon and rule-based approach to sentiment analysis, leveraging a pre-built dictionary of words and their associated sentiment scores. Table 4 demonstrates a subset of the VADER dictionary.

The following provides a simplified overview of VADER's functioning:

1. Lexicon-based approach: VADER utilizes a lexicon or dictionary containing a comprehensive collection of words and their sentiment intensity scores. Each word in the lexicon is assigned a score indicating its positivity, negativity, or neutrality.
2. Polarity scoring: VADER evaluates the sentiment of a given text by computing a sentiment polarity score. This process involves analyzing both individual words and phrases in the text.
3. Valence scoring: VADER considers the valence of individual words, representing the degree of positive or negative sentiment associated with each word. The lexicon includes intensity modifiers like "very" or "extremely" to adjust the valence of preceding words.
4. Sentiment aggregation: VADER combines the valence scores of all words in the text, taking into account grammatical and syntactical structures. It considers the context in which words appear and applies grammatical rules to handle negations, punctuation, and capitalization.
5. Normalization: To generate a sentiment polarity score, VADER normalizes the aggregated sentiment score within a range between -1 and +1. A score close to +1 indicates strong positive sentiment, while a score near -1 represents strong negative sentiment. A score around 0 suggests neutral sentiment.
6. Sentiment classification: Based on the final sentiment polarity score, VADER classifies the text as positive, negative, or neutral. Additionally, VADER provides a compound score that represents the overall sentiment intensity, combining all the sentiment scores.

VADER's strength lies in its ability to handle sentiment analysis for informal and colloquial text, including social media posts, online reviews, and informal language usage. It is specifically designed to handle sentiments expressed in contexts where traditional sentiment analysis approaches may struggle.

Using VADER in sentiment analysis provides deeper insights into the sentiment expressed in reviews and enables more effective analysis of customer feedback. The sentiment scores and categories provided by VADER will be added as additional columns to the final dataset, facilitating further analysis of the Airbnb listings.

Once the sentiment analysis has been completed, the remaining missing values in the dataset can be filled by using the average sentiment scores associated with the respective host id numbers. This ensures that the missing values are replaced with sentiment scores that accurately represent the sentiment expressed by each specific host. By utilizing the average sentiment scores from the corresponding host id, the imputed values offer a reasonable estimation of the sentiment for the unavailable data points. Three new columns are created, these columns are: year, month, and day of the week. These new columns will become necessary when merging the calendar data.

After addressing the missing values the calendar data and reviews data are combined with an inner join. The calendar data provides supplementary contextual details such as price, date, month, year, and day of the week, which can be correlated with the review sentiments. By merging the two datasets based on shared columns like month, year, day of the week, and listing id, a comprehensive dataset is created that consolidates information from both reviews and calendar data.

Table 4: Subset of sentiment ratings for words in the VADER dictionary.

| Word | Sentiment Rating |
|---|---|
| Tragedy | -3.4 |
| Insane | -1.7 |
| Flattery | 0.4 |
| Stealthily | 0.1 |
| Awesome | 1.8 |
| Amazing | 1.8 |

## 4.2 DATA EXPLORATION

This section presents the findings of an in depth analysis conducted on Airbnb price data with the primary objective of gaining insights into the distribution, normality, and seasonal variations of Airbnb features. Understanding these key aspects of the data is crucial for making informed decisions in the Airbnb industry, such as setting competitive prices, optimizing occupancy rates, and predicting demand patterns.

### 4.2.1 EXPLORING LISTINGS

The listings data is presented all in one data frame. Its column features provide descriptive information, and exploration was performed to measure the importance of the features ahead of preprocessing. There is no time-series aspect towards the listings data frame which removes the need to search for trends seen across the data. That being said, exploring the prevalence and abundance of descriptive columns for each listing was necessary. Additionally, exploring whether or not a column was of interest or if it is to be dropped from the analysis was required. To perform this task, thought and discussion were put into each column with the group over if a column was integral for the model.

To realize how many room types there were around Austin, a second bar chart was created as shown in Figure 2.

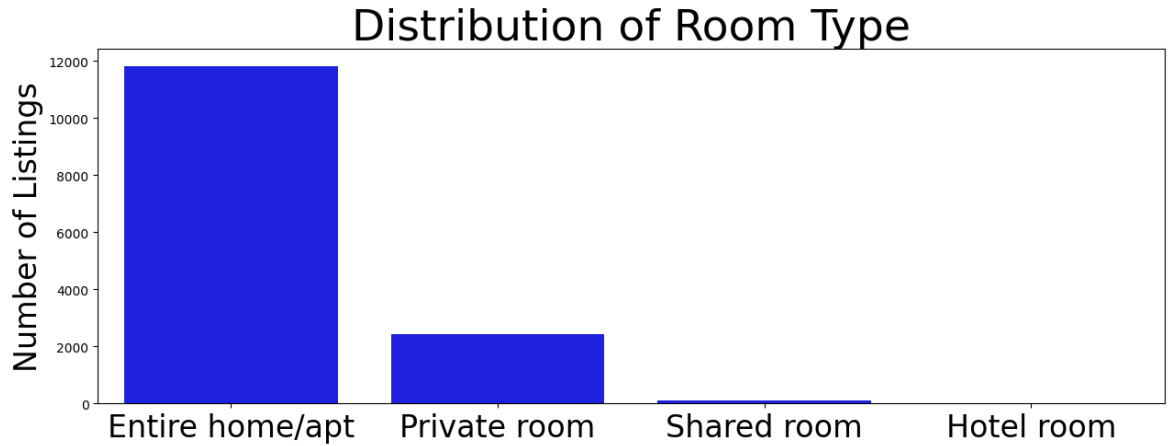

Figure 2: Bar chart based on the type of room in the listings dataset.

As seen in Figure 2, it was clear that the majority of room types were either entire homes or entire apartments; although, there were a significant amount of private rooms prevalent in the dataset. Then to realize the concentration of property types, a bar chart was created on the 'property_type' column as shown in Figure 3. This figure displays the top ten of the seventy different property types on the dataset. As seen in the figure, there are a diverse range of property types, but their popularity tails off significantly quickly.

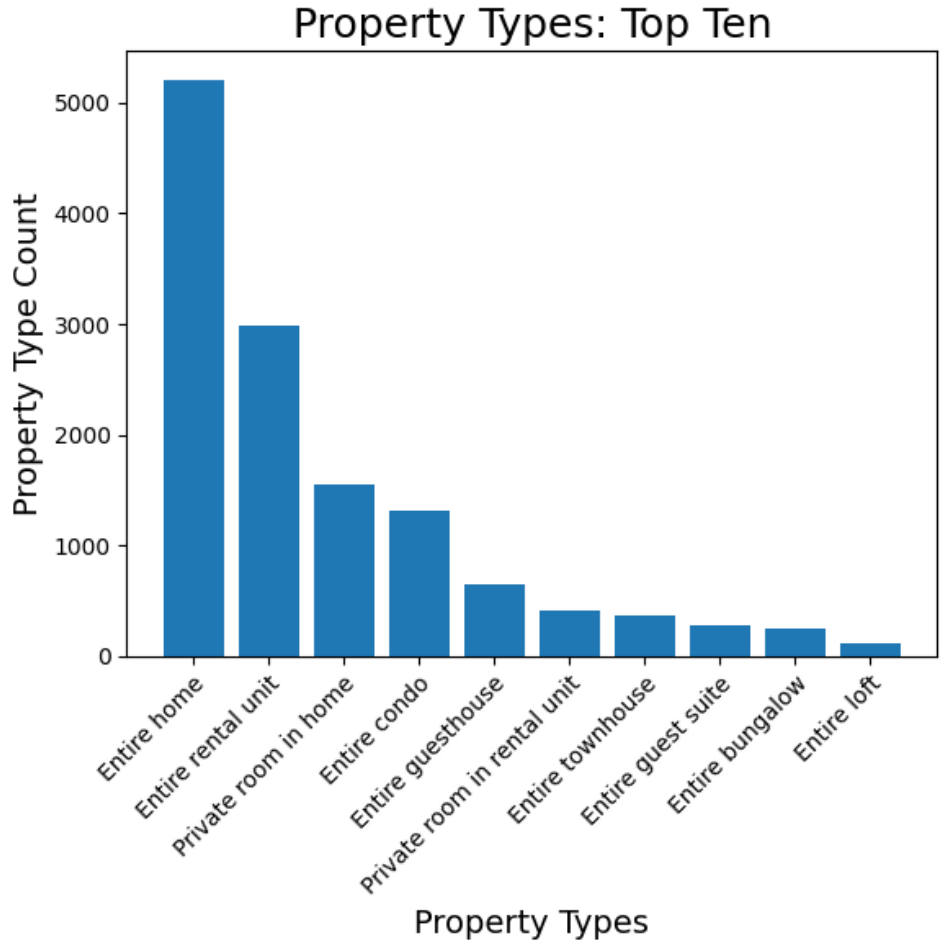

Figure 3: A bar chart showcasing the distribution of property types in the listings dataframe. The chart highlights the ten most commonly occurring property types.

In Figure 4, a bar chart has been created to display the frequency of listings within the ten most densely populated zip codes. The group's interest was in understanding the importance of location for each listing by analyzing the number of listings per zip code..

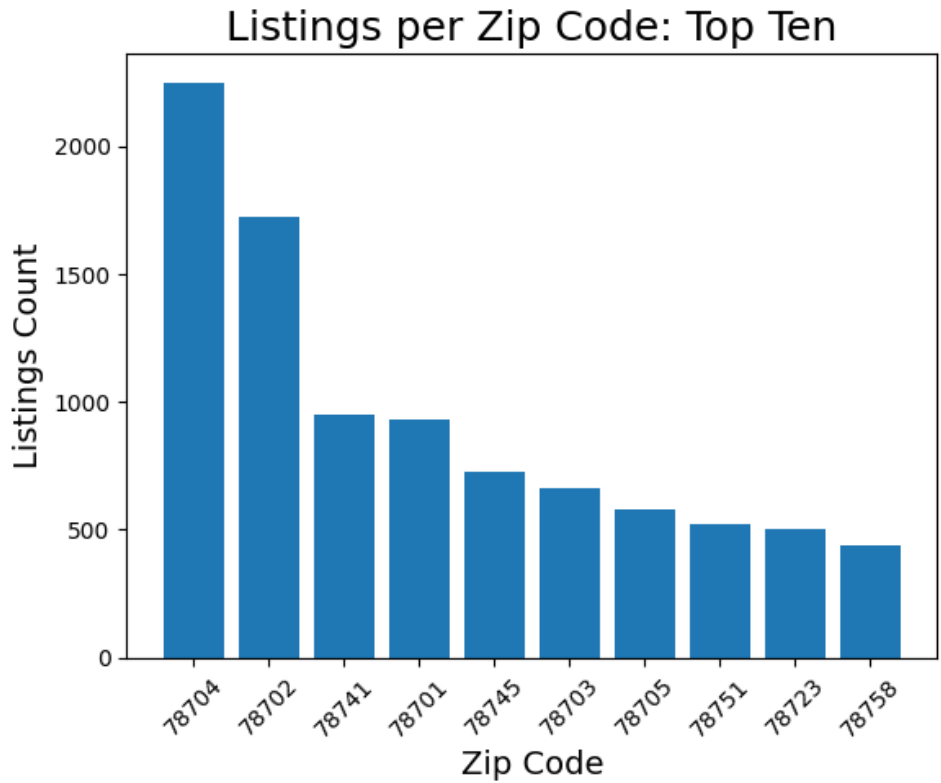

Figure 4: Top ten zip codes for the amount of listings within Austin, TX.

Figure 5 showcases the distribution of listings across different zip codes, providing insights into how concentration varies across Austin. To gain a clearer understanding of location-related price differences, Figure 6 and Figure 7 were generated. These figures depict the impact of location on pricing and correlate with the concentration of listings presented in Figure 5.

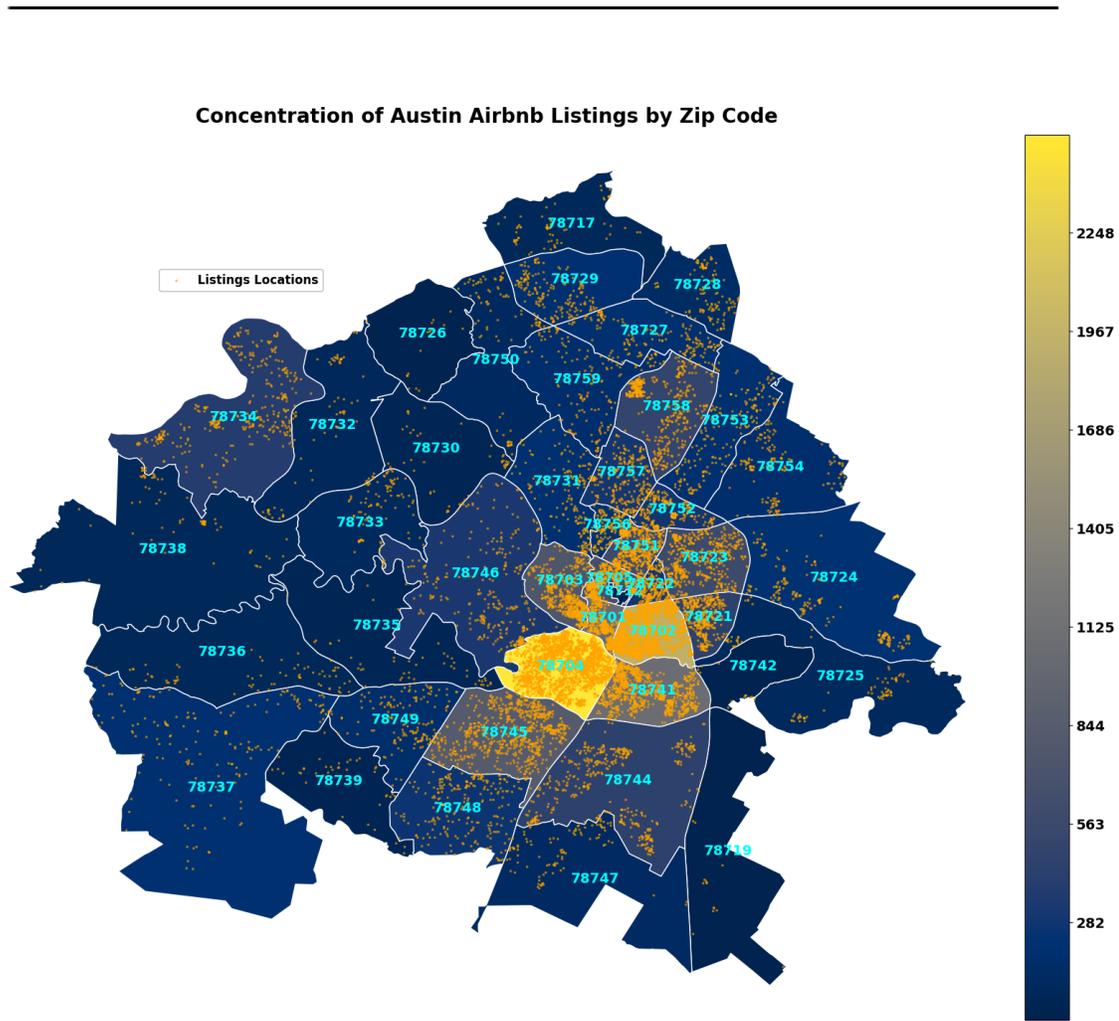

Figure 5: Concentration of listings per zip code as a gradient with Austin. Markers represent listings locations.

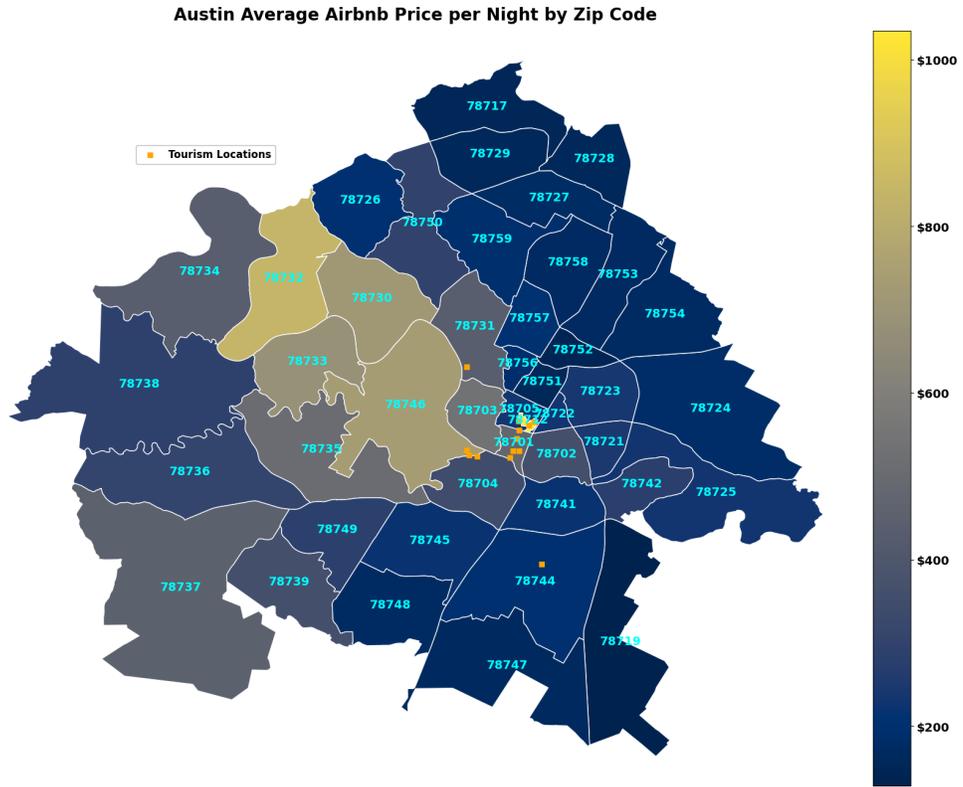

Figure 6: The average price of every Airbnb listing within a zip code as a color gradient across Austin. Markers shown for the top 13 most popular tourism locations.

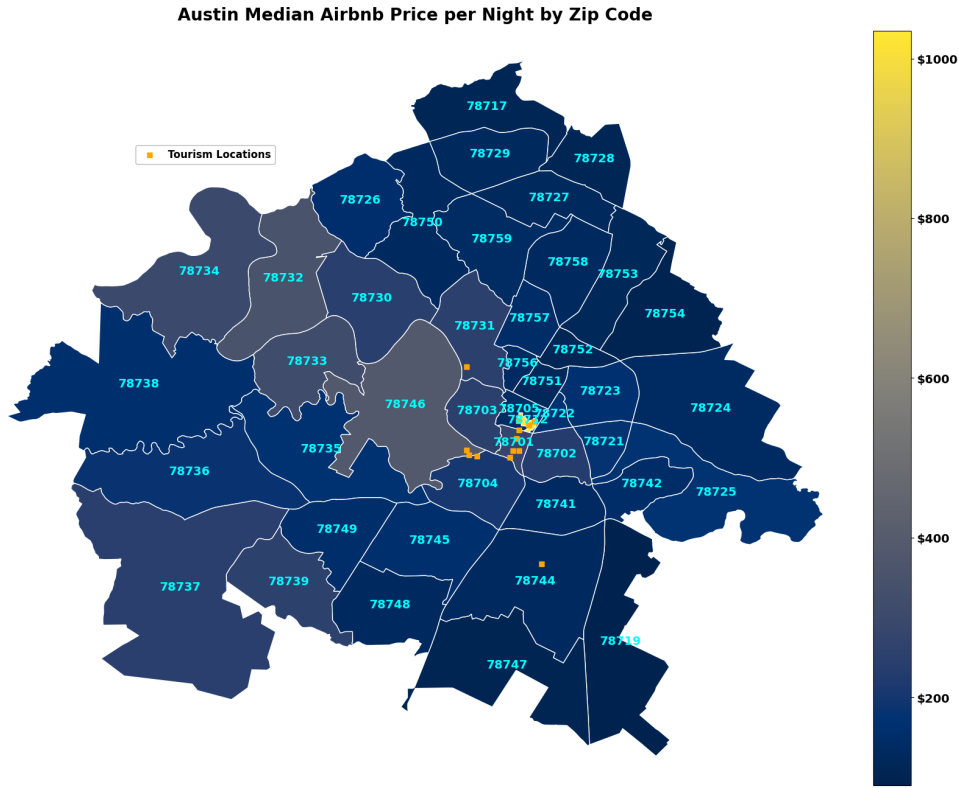

Figure 7: The median price of every Airbnb listing within a zip code as a color gradient across Austin. Markers shown for the top 13 most popular tourism locations.

Figures 5, 6, and 7 give valuable insight as to how location across Austin is impactful. There is a steady gradient of Airbnbs per each zip code outside of what is known as central Austin. Additionally, the pricing of Airbnbs drops off outside of the central area. Seen in Figure 6 and compared with Figure 7, there are several outliers that represent higher prices per night towards the western to northwestern zip codes of Austin, TX. Those higher average priced zip codes overlap with a more expensive area of Austin and border Lake Travis. That being said, the price value listed on the listings data frame represents the price for the night of stay at each Airbnb location on the date the data was scraped from Airbnb. There are two different dates for scraping shown in the raw import and the listings differ between the two - March 16$^{th}$ and 17$^{th}$, 2023. It is worth noting that those dates span a Thursday and a Friday. As seen from exploring the calendar data, many listings differ in price between those two days of the week.

Following the visualization side of exploration, the data was transferred into a pandas data frame and simple counts were carried out to realize the diversity across columns of interest. Finding out what these columns represented was necessary for some, but not all. The majority of columns are described in full in an additional document provided by Inside Airbnb that is shown in Table 9.

### 4.2.2 EXPLORING CALENDAR

Data was collected from the historical data of all Airbnb listings, capturing the average price for each month of the year as well as day of the week. A boxplot and line plot were generated to visualize the different price trends.

Boxplots are a useful tool for analyzing Airbnb prices as it provides insights into the central tendency, spread, and skewness of the price distribution, as can be seen on Figure 8. Analyzing the box plot of Airbnb prices yielded interesting findings. When considering the days of the week from Sunday to Thursday, the median price (Q2) was approximately $155, indicating that half of the listings had prices below this value. The IQR, represented by the box, spanned from around $100 (Q1) to $260 (Q3). On Fridays and Saturdays, the median price slightly increased to around $180, while the IQR expanded from $100 to $290. The IQR encompasses the middle 50% of the data, highlighting the typical price range for most listings.

The seasonal pattern analysis revealed two prominent peaks in median prices, with the first peak occurring in March and the second peak in October. Figure 9 visually represents the monthly fluctuations of Airbnb prices, providing a clear depiction of the observed trends and seasonal variations. The first peak in March indicates a potential correlation with seasonal trends, likely due to factors such as school breaks, vacations, and favorable weather conditions. Events like South by Southwest may also contribute to the increased demand and subsequently higher prices during this time. Similarly, the second peak in October suggests an association with events like the Austin City Limits Festival, cultural activities, and Halloween festivities. The increased demand during this time leads to higher prices as hosts capitalize on the seasonal attractions and festivities. Overall, these findings indicate that both March and October are popular periods for Airbnb bookings due to increased demand and potential events occurring during those months. Understanding these peak months is crucial for hosts and travelers. Hosts can optimize their pricing strategies to maximize revenue during these high demand periods, while travelers can plan their trips accordingly, taking into account the potential higher prices and availability constraints during these months.

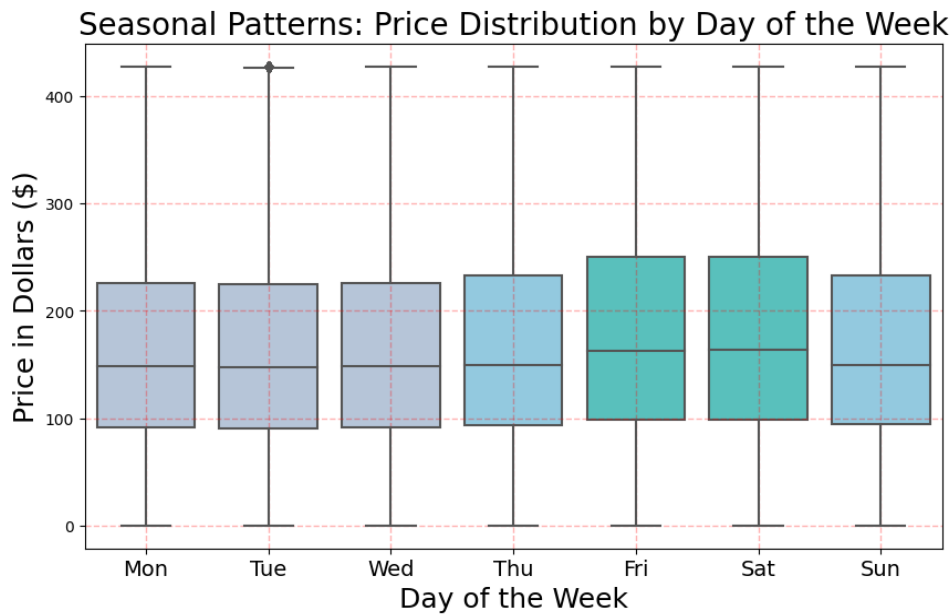

Figure 8: Variations of Airbnb prices by day of the week. The boxplot showcases the price variation of Airbnb listings based on different days of the week, with median prices of around $155 (Sunday to Thursday) and $180 (Friday and Saturday), highlighting the typical price ranges for each category.

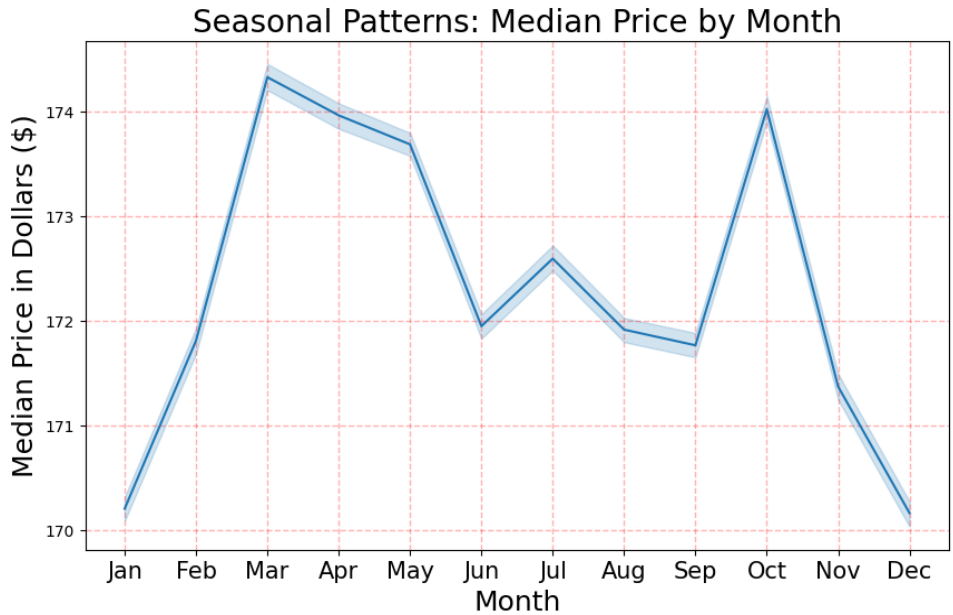

Figure 9: Monthly fluctuations based on median Airbnb listing prices.

#### 4.2.3 EXPLORING REVIEWS

In today's digital age, online reviews have become an integral part of the consumer decision making process. Before making a purchase, booking a service, or choosing a destination, individuals often turn to online platforms to read reviews and gather insights from the experiences of others. Exploring reviews data can uncover patterns, sentiments, and correlations that offer deep insights into customer satisfaction, areas for improvement, and overall brand perception. As a result, reviews data has become a valuable resource for businesses, and consumers alike.

Histograms are a powerful tool for understanding the distribution of data, allowing us to identify trends and patterns that may be hidden in raw numbers. In this case, the histogram in Figure 10 displays a left-skewed distribution, indicating a notable variation in the frequency of reviews across the years. Towards the left end of the histogram, corresponding to the earlier years such as 2009, the number of reviews were significantly lower. This suggests that during the early years, there was relatively less activity or engagement in the Airbnb platform. As we move towards the right end of the histogram, representing the more recent years, there is a noticeable increase in the number of reviews. This surge in review activity suggests a higher level of user engagement and a growing emphasis on providing feedback in the more recent years.

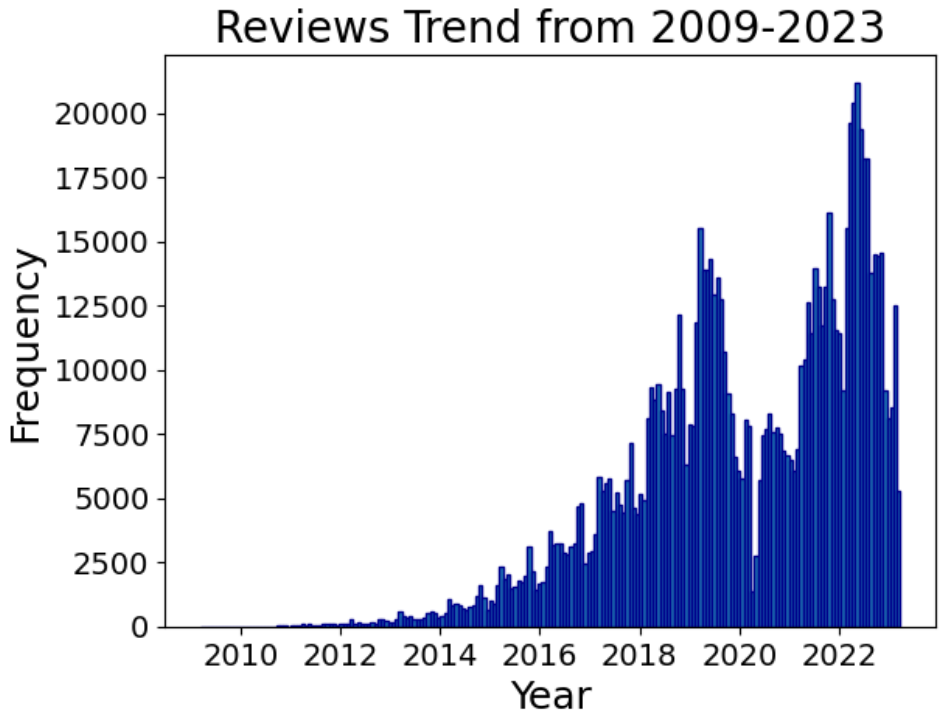

Figure 10: Histogram of Airbnb reviews from 2009 to 2023.

Another valuable aspect of exploring Airbnb reviews data is gaining insights from textual analysis, such as word clouds. Word clouds visually represent the most frequently used words in the reviews left for Airbnb listings. By examining the prominent words that stand out in the word cloud, hosts can gain a deeper understanding of the key themes and sentiments expressed by guests.

The word cloud in Figure 11, depicts the prominent terms used in Airbnb reviews. The larger the word appears, the more frequently it is mentioned in the reviews. The word cloud highlights significant terms such as "highly recommend," "great location,", "walking distance,", and "everything needed." These terms shed light on the attributes and experiences that guests frequently emphasize and appreciate in their Airbnb stays.

Incorporating the insights gained from the Airbnb reviews word cloud into the overall analysis provides a more comprehensive understanding of customer sentiments and preferences. With combining the findings from the histogram and the word cloud, businesses can identify trends in review activity over time, while also understanding the specific features and experiences that drive positive guest reviews.

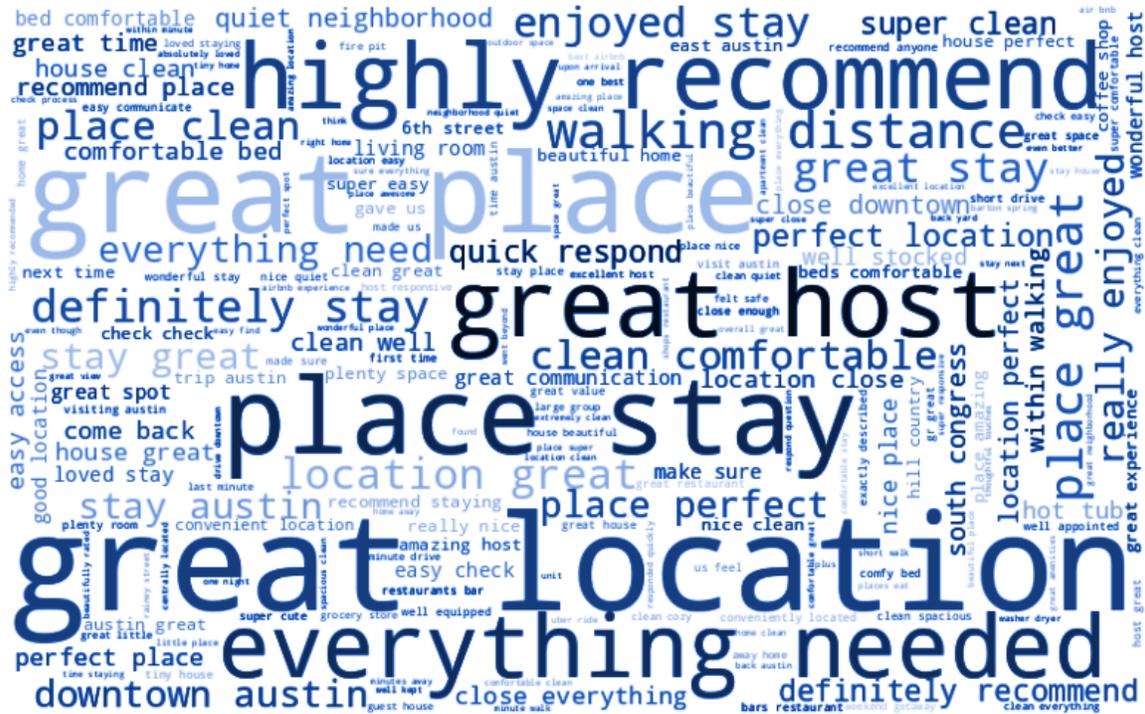

Figure 11: Word Cloud of prominent Airbnb review terms

4.3 DATA MODELING

Based on our literature review and subsequent analysis, this data science project focuses on predicting Airbnb prices. To achieve this objective, five machine learning models will be utilized: Lasso regression, Ridge regression, ElasticNet, Random Forest regression, and XGBoost.

Linear regression models (Lasso, Ridge, and ElasticNet) are commonly employed in price prediction tasks and will serve as a good baseline for our project. Previous studies, such as Yu et al. (2018), have demonstrated the effectiveness of linear regression in predicting real estate prices. By considering relevant features such as location, property size, amenities, and other factors, linear regression can capture the linear relationship between these features and the target variable (price). It is important to note that linear regression assumes a linear relationship, which may not fully capture the complexity of the Airbnb pricing dynamics.

Random forest regression, on the other hand, is a powerful ensemble learning technique that can handle non-linear relationships and complex interactions among features. Random forest regression has been successful in predicting prices in the Boston housing market. Adapting this approach to short-term rental values presents unique challenges. Short-term rental prices tend to be more dynamic and subject to frequent fluctuations. Additional considerations and modifications will be required to apply random forest regression effectively in this context.

XGBoost, a gradient boosting algorithm, stands out as a robust choice for predicting Airbnb prices. It excels in handling complex relationships within data and has demonstrated exceptional performance in various domains, including regression tasks. XGBoost constructs an ensemble of decision trees, sequentially refining the model's predictions by emphasizing the patterns that were previously difficult to capture. Its ability to handle non-linearity, feature interactions, and feature importance ranking makes it a suitable candidate for our prediction task.

In our project, we have identified certain constraints derived from the data wrangling process, and data explorations. These constraints include:

- Data Quality: The presence of missing values or outliers in the dataset required appropriate handling strategies such as imputation and outlier removal.
- Categorical Variables: In our dataset, some categorical variables required appropriate encoding and feature engineering techniques to represent these variables numerically.

- Data Normalization: Some models benefit from data normalization to ensure features are on a similar scale and avoid bias towards certain features data normalization should be performed.
- Computational Resources: The computational requirements of certain models should be considered to ensure feasibility within the available resources.

To address the generalization problem and further enhance the performance of these models, the following techniques can be implemented:

1. Feature Engineering: This involves creating new features or modifying existing ones to improve model performance. This could include things like encoding categorical variables, normalizing numerical features, or creating interaction terms.

2. Regularization: Regularization is a technique that adds a penalty term to the model's objective function, discouraging overly complex models. Common forms of regularization include L1 regularization (Lasso) and L2 regularization (Ridge). By imposing constraints on the model's parameters, regularization helps control overfitting and improves generalization.

Interpretability is an important aspect of any machine learning project. In this project, interpretability refers to the ability to understand and explain how the chosen machine learning models derive their final predictions and identify which features contribute to influencing Airbnb prices. Models like linear regression provide inherent interpretability through coefficients assigned to each feature. Other models like random forest regression may require additional techniques to enhance interpretability.

To enhance interpretability, the following approaches will be implemented:

1. Feature Importance Analysis: Feature importance analysis helps identify the relative importance of each feature in the model's predictions. For linear regression, this can be achieved by examining the coefficients associated with each feature. In random forest regression, feature importance can be determined based on the average impurity decrease caused by each feature across all decision trees in the forest.

2. SHAP (SHapley Additive exPlanations) values: It calculates the contribution of each feature to the prediction for each instance. Unlike partial dependence plots, SHAP takes into account interactions between features and gives a more accurate depiction of the effect of each feature.

With applying feature selection and regularization, we can mitigate the generalization problem and improve the models' ability to generalize to unseen data. We can also employ techniques such as feature importance analysis and SHAP values to enhance the interpretability of the selected models, which can allow us to gain insights into the factors contributing to Airbnb price predictions.

5. EXPERIMENTS

Feature selection and train-test split are fundamental steps in the machine learning pipeline that play a crucial role in model development, performance evaluation, and generalization to unseen data.In this section, we discuss the significance of feature selection and the train-test split process in our research on predicting Airbnb prices.

5.1 SETUP

5.1.1 FEATURE SELECTION

Feature Selection was found to be important for the model during the early stages of review and exploration of the data. The initial count of features was robust, and the preprocessing expanded them even further to a larger count. In order to reduce the amount of features and thus reduce noise in the model, feature selection was implemented. That being said, several trials were carried out to assess the positive impact of feature selection and compare different methods throughout the course of the project.

SelectKBest was the first trial. SelectKBest is a class provided by the scikit-learn library that is used for feature selection in machine learning. Feature selection is the process of selecting a subset of relevant features (also known as variables or attributes) from a larger set of features available in a dataset. The goal is to choose the most informative and discriminative features that are likely to have a significant impact on the predictive performance of a machine learning model. SelectKBest is particularly useful when dealing with high-dimensional datasets where the number of features is large. It helps to reduce the dimensionality of the data by selecting the top K features based on their importance scores.

Feature selection is important for several reasons:
- Improved Model Performance: By selecting the most relevant features, the model can focus on the most informative aspects of the data, potentially leading to better prediction accuracy and generalization.
- Reduced Overfitting: When dealing with high-dimensional datasets, models can become overly complex and prone to overfitting. Feature selection helps to mitigate this problem by eliminating irrelevant or redundant features, reducing the complexity of the model.
- Enhanced Interpretability: Feature selection allows for a more interpretable model by identifying the subset of features that have the strongest relationship with the target variable. This can provide valuable insights and help understand the underlying factors driving the predictions.

For this research, we utilized the SelectKBest method with a scoring function designed for classification tasks. The specific scoring function we used is associated with the ANOVA (Analysis of Variance) F-value. This F-value assesses the linear relationship between a feature and the target variable (in this context, the price). It measures the variance between different categories of the target variable and compares it to the variance within each category. The F-value indicates the importance of the feature concerning the target variable. The scoring function computes both the F-value and corresponding p-values for each feature. A higher F-value indicates a more pronounced relationship between the feature and the target variable. The SelectKBest approach ranks features according to their F-values and selects the top K features with the highest scores.

It's worth noting that the scoring function assumes normally distributed features and equal variances across different categories of the target variable. If these assumptions are not met, alternative scoring functions such as mutual information or the chi-squared test can be considered. Mutual information measures the statistical dependence between two random variables, indicating the information obtained about the target variable from a specific feature. The chi-squared test calculates the chi-squared statistic and p-value to determine the significance of the relationship between two categorical variables.

In our research, the SelectKBest method was applied to the dataset, using K = 40 (selecting 40 features), with the price variable as the target variable. The following results were obtained:

Table 5: Top 40 feature selection scores.

| Feature | Score |
|---|---|
| room_type_Private room | 1310.3853 |
| bedrooms | 1261.6139 |
| accommodates | 1183.7553 |
| room_type_Shared room | 1044.0561 |
| beds | 997.1433 |
| reviews_per_month | 331.3587 |
| Dishwasher | 307.2222 |
| Kitchen | 270.1310 |
| number_of_reviews_ltm | 257.0923 |
| Number of Amenities | 214.9709 |
| number_of_reviews | 206.3285 |
| Cooking basics | 192.7666 |
| number_of_reviews_l30d | 189.5228 |
| property_type_num | 177.2615 |
| Private entrance | 151.9356 |
| Dishes and silverware | 133.6048 |
| Hangers | 120.8480 |
| review_scores_rating | 118.2396 |
| Smoke alarm | 111.8495 |
| availability_30 | 111.0043 |
| review_scores_cleanliness | 107.5314 |
| Self check-in | 102.5307 |
| host_is_superhost | 102.2706 |
| availability_365 | 100.6833 |
| instant_bookable | 100.6772 |
| year | 99.9961 |
| Oven | 99.0230 |
| zilker | 97.1199 |
| mex_arte | 96.5390 |
| congress_bridge | 96.3372 |
| umlauf | 96.2588 |
| barton_springs | 96.1993 |
| Carbon monoxide alarm | 96.0981 |
| museum_weird | 96.0288 |

| Hot water | 95.2546 |
| t_cap | 95.0391 |
| host_since | 94.5443 |
| bullock | 92.9782 |
| availability_60 | 91.1884 |
| ut_tower | 90.7459 |

While impactful, more modern approaches to feature selection underwent trial. These more modern approaches consisted of two methods: forward selection and SHAP.

Forward selection is a stepwise feature selection technique that constructs a predictive model by iteratively adding the most relevant features to the model. The process starts with an empty model and progressively incorporates one feature at a time based on their contribution to the model's performance. In our implementation, Mean squared error, or MSE, was used to evaluate the performance of a feature in a linear regression model. Once all the features had been measured, the best scoring feature was selected and appended to a list. From there, the process started over selecting the next best feature in accordance with the feature on the list, and later on in conjunction with the features that have been added previously. This process carries on until a predetermined measure is hit for whether or not a feature is 'important' or helps the model to a certain degree.

Forward selection is thought to be a more suitable option for several reasons. Unlike SelectKBest, forward selection has an approach that seems more in tune with the goal for this project's model. It is adaptive meaning that features are chosen based on their ability to enhance the model's performance. SelectKBest, by contrast, selects features based on precomputed scores without considering their potential interaction effects or impact on the final model. Additionally, forward selection builds the feature subset step by step, evaluating the contribution of each feature in the context of the current model. This sequential nature allows it to potentially capture interactions between features that SelectKBest might miss since the latter only considers individual feature scores. Lastly, SelectKBest relies on predefined statistical metrics to rank and select features. This can introduce a bias towards features with high scores, regardless of their actual impact on model performance. Forward selection, on the other hand, assesses features within the context of the evolving model's performance.

Forward selection resulted with a list of 85 features that were important for the model and built progressively off of one another. Nonetheless, another modern technique for feature selection was executed to judge the results and see which had a higher impact.

SHapley Additive exPlanations, or better known as SHAP, is a powerful technique for explaining the predictions of machine learning models, and it can also be used for feature selection. Its design is rooted in cooperative game theory and provides a principled way to attribute the contribution of each feature to the model's predictions. The technique first quantifies the individual contribution of each feature to a specific prediction made by a model. All possible combinations of features are considered as well as the average contribution of a feature when it's included or excluded from these combinations. Baseline predictions are compared to the actual prediction to determine how features shift the prediction. Importance of features is ranked which contributes to selection. SHAP values satisfy desirable properties like local accuracy, the sum of SHAP values equals the prediction, and consistency meaning if a feature is less important for all instances, its SHAP value is lower for all instances. This makes SHAP suitable for sensitive tasks like fairness analysis.

Overall, SHAP offers a rigorous and interpretable way to evaluate the importance of features in a model's predictions, which can guide feature selection decisions by identifying the most influential variables for the task at hand. SHAP results were taken for both Linear Regression and Random Forest models to judge its impact. The results are seen in section 5.3.

SHAP has strengths where forward selection lacks. Forward selection is known to overfit without careful implementation and is sensitive to feature additions. That being said, forward selection was deemed the more desirable choice due to its speed, simplicity, and keen ability to work with the dataset size of the project and was implemented for the modeling section. Although, SHAP still added value to the overall project by adding in supplemental information and visuals towards feature importance as mentioned in section 4.3 and results shown in section 5.3.

5.2  EXPERIMENTAL RESULTS

**Lasso & Ridge Regression Results:**
With the L1 regularized (Lasso) regression and the L2 regularized (Ridge) regression, the performance of the two models are extremely similar regardless of forward selection being applied or not. As seen in Table 6, without forward selection we saw a near identical r-squared and very minute differences in their MAE and RMSE. In Table 7 with forward selection being applied we see the ridge regression perform slightly better than the lasso regression, with the ridge getting an r-squared of about 0.52 and the lasso regression seeing an r-squared of approximately 0.51. This could potentially be due to the fact that ridge regression's regularization can better handle multicollinearity, allowing it to marginally outperform Lasso regression in our price prediction scenario.

**ElasticNet Results:**
Since ElasticNet is the combination of both L1 and L2 regularization, it is interesting to observe its performance be about the same if not slightly worse than the lasso or ridge regressions individually. With the non-feature selected results it can be seen that the ElasticNet performed about on par with the lasso and ridge regressions. However with the forward selection applied, the ElasticNet performed worse than both the lasso and ridge regressions. In regards to feature selection being applied, the ElasticNet provided an r-squared of 0.48 which is lower than both lasso and ridge, and had higher error scores than both. ElasticNet might perform similarly or slightly worse than lasso or ridge regression due to the fact that our dataset potentially has more clear characteristics that align with the strengths of lasso or ridge individually. Combining both penalties in ElasticNet could introduce unnecessary complexity, leading to comparable or slightly worse performance.

**Random Forest Regression Results:**
As seen in Table 6, the Random Forest regression performed the best in predicting the daily listing price, with an MAE of 29.8, an RMSE of 48.3, and an r-squared of 0.79. Furthermore, the feature selected results proved to be even better. With an r-squared of 0.89, MAE of 16.1, and an RMSE of 35.4. This is most likely due to the fact that the Random Forest Regression model performs well because of its ability to handle complex interactions and capture non-linear relationships in the data. It leverages the ensemble of decision trees, averaging their predictions to reduce overfitting and improve generalization. The model's robustness to outliers and missing values also contributes to its superior performance in predicting the daily listing price. And since this was our best performing model, we will fine tune the hyperparameters of this model to enhance accuracy and or run-time.

**XGBoost Regression Results:**
XGBoost is a powerful algorithm for predicting daily listing prices due to its ability to handle complex relationships, non-linearities, and high-dimensional feature spaces. It incorporates boosting, which sequentially adds decision trees to correct errors made by previous trees. Without feature selection it performed quite well with an RMSE of 51.4, MAE of 35.1, and an r-squared of 0.77. And these performance scores were relatively the same even with feature selection applied. However, despite its strengths, XGBoost sometimes falls short of outperforming Random Forest Regression in predicting daily listing prices. One reason is that XGBoost can be more prone to overfitting compared to Random Forest. With its iterative nature, XGBoost is susceptible to memorizing noise or outliers in the training data, leading to reduced generalization performance. Random Forest, on the other hand, constructs individual decision trees independently and combines their predictions through averaging, providing a built-in mechanism to mitigate overfitting.

Overall, our ensemble learning techniques did outperform all three sets of linear regression which were mainly used as reference models. Our best performing technique was the Random Forest Regression model, with it having the lowest RMSE and MAE and having the highest r-squared. This model achieves the goal of being able to predict the daily listing within a reasonable amount, with interpreting the MAE of being approximately $16.1 (feature selection applied). So being able to capture a listing price of within $16.1 could be extremely useful for both hosts and guests on Airbnb.

Table 6: Model results without Forward-Selection.

| Metric | Lasso | Ridge | Elastic | Random Forest | XG-Boost |
|---|---|---|---|---|---|
| R-Squared | 48% | 48% | 48% | 79% | 77% |
| Mean Absolute Error | 56.7 | 56.6 | 57.1 | 29.8 | 35.1 |
| Root Mean Squared Error | 76.6 | 76.4 | 76.9 | 48.3 | 51.4 |

Table 7: Model results with Forward Selection.

| Forward Selection | | | | | |
|---|---|---|---|---|---|
| Metric | Lasso | Ridge | Elastic | Random Forest | XG-Boost |
| R-Squared | 51% | 52% | 48% | 89% | 77% |
| Mean Absolute Error | 54.7 | 54.3 | 57.0 | 16.1 | 35.5 |
| Root Mean Squared Error | 74.5 | 74.2 | 77.0 | 35.4 | 51.8 |

5.3 CROSS- VALIDATION RESULTS

Hyperparameter tuning is a critical step in building robust machine learning models. It involves finding the best set of hyperparameters that maximize the performance of the model. We utilized Random Search Cross-Validation to identify optimal hyperparameters for our regression models. The dataset used for training and evaluation consists of a scaled version of the feed-feature selected historical Airbnb data.

To facilitate the hyperparameter tuning process, specific hyperparameter grids were defined for each regressor. These grids encompass a range of hyperparameter values that will be explored during the Random Search process. The hyperparameters under consideration include regularization strengths, alpha values, and various parameters related to the ensemble methods.

For this project, a 5-fold cross-validation approach is utilized, dividing the dataset into five equally sized subsets for training and validation. Within each fold, Random Search is employed to randomly sample hyperparameter combinations from the defined grids, exploring the hyperparameter space.

The Random Search Cross-Validation process successfully identified optimal hyperparameters for each regression model in the context of predicting Airbnb prices. Among the considered models, Random Forest Regression emerges as the top, achieving a score of approximately 80%. This suggests that Random Forest Regression is best suited to capture the complexity of Airbnb price prediction. While Random Forest Regression excels, it did not exhibit a significant improvement in performance compared to the results obtained without cross-validation. This variance in performance could be attributed to a combination of factors including the dataset's characteristics, the models' inherent strengths, and the intricacies of the cross-validation process itself. The remaining models did not exhibit significant improvement compared to the results obtained without cross-validation.

### 5.4 SHAP ANALYSIS

To enhance the interpretability of our best performing model, Random Forest, a SHAP analysis was conducted. SHAP values allow us to quantify the impact of each feature on the model's prediction for individual instances, providing a better understanding of feature importance both globally (across the dataset) and locally (for individual predictions).

From the SHAP analysis in conducted on the Airbnb price prediction model, illustrated in Figure 12, the top three features that stood out in terms of impact on the model's prediction are:

1. Bedroom: The amount of bedrooms in an Airbnb listing significantly influences the predicted price.
2. Congress Bridge: Indicating whether the listing is in close proximity to Congress Bridge or not has a substantial impact on the predicted price.
3. Day of the Week: Price variations based on different days of the week play a crucial role in determining the predicted price.

These features possess the highest mean absolute SHAP values, signifying their substantial impact on the model's decisions across the entire dataset. The blue colored bars in the SHAP plot indicate the features' effects on predictions. The predominance of blue suggests the extent to which these features influence the predictions.

While this SHAP analysis provides valuable insights into the Random Forest Regression model's decision-making process, it's important to note that the presence of influence doesn't imply direct causality.

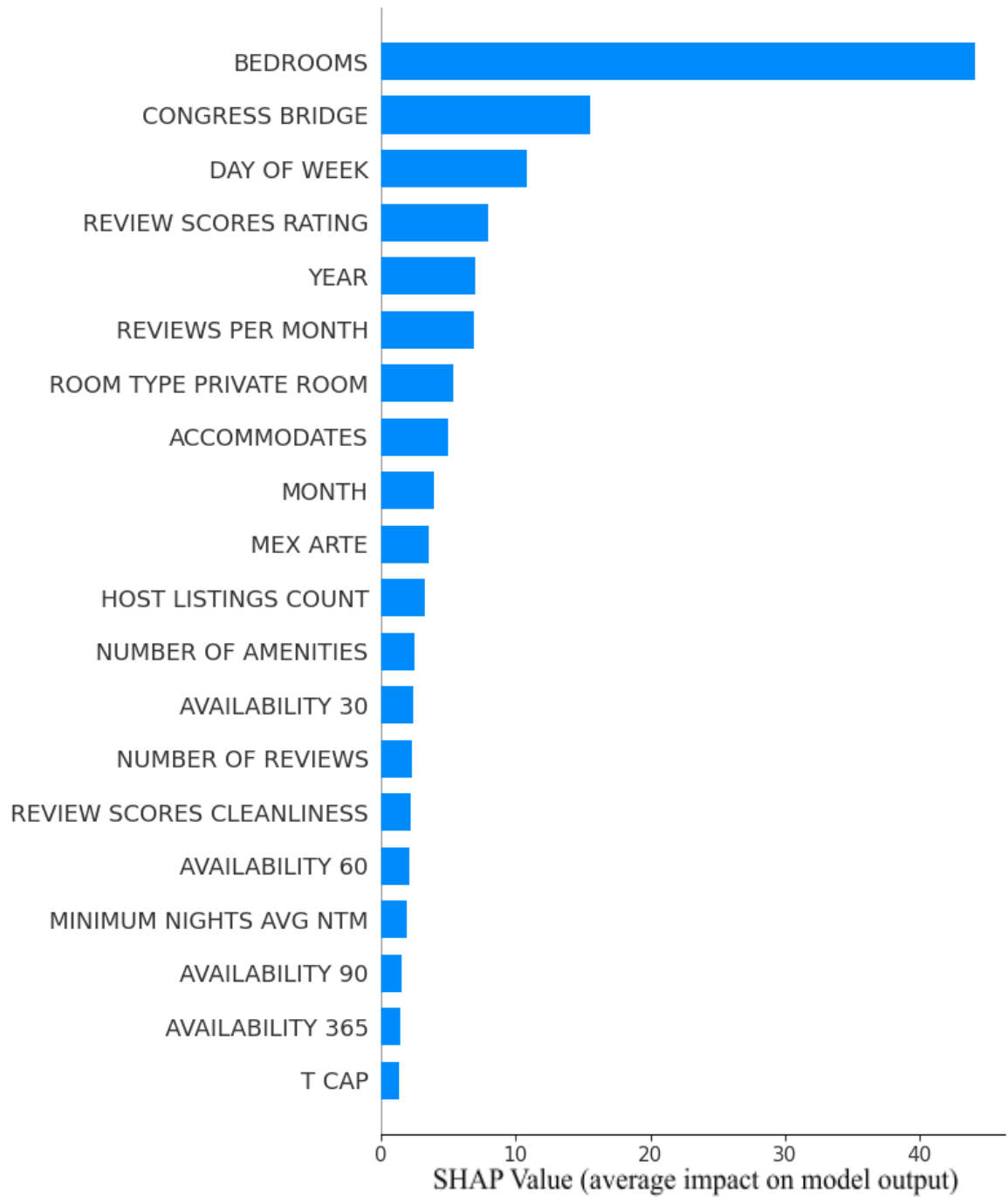

Figure 12: Random Forest feature impact analysis: top 20 influential features revealed by SHAP.

6. CONCLUSION

Our research aimed at predicting Airbnb listing prices using regression models has provided valuable insights into addressing a significant challenge in the hospitality industry. Through careful data pre-processing, strategic feature engineering, and employing a range of regression techniques, we've constructed a data science framework for predicting Airbnb listing prices.

Our analysis has highlighted the effectiveness of various regression models, including Linear Regression with L1, L2, and Elastic Net regularization, Random Forest, and XGBoost. These models were carefully selected and fine-tuned to ensure optimal performance in predicting listing prices accurately. Notably, the Random Forest model emerged as the best-performing model in this context, consistently demonstrating superior predictive capabilities.

We utilized a set of evaluation metrics, encompassing accuracy, Mean Squared Error (MSE), Root Mean Squared Error (RMSE), and Mean Absolute Error (MAE). These metrics provide a clear view of our models' performance, enabling us to assess their accuracy, and overall predictive capabilities.

Our SHAP analysis has brought attention to the critical role played by certain key features in influencing Airbnb listing price predictions. This analysis not only enhances the interpretability of our models but also equips hosts and guests alike with valuable insights into making informed decisions related to pricing strategies and customer satisfaction.

As we progress, we acknowledge the potential for further improvement. Future efforts could focus on addressing limitations, enhancing model interpretability even further, and delving into the potential impact of external variables on listing prices. These avenues hold the promise of elevating our predictive framework and maximizing its practical relevance in the dynamic world of Airbnb pricing.

## 6.1 IMPACT

Our data science pipeline combines advanced techniques such as machine learning modeling, feature engineering, and sentiment analysis to accurately predict Airbnb prices. Leveraging historical data, we have developed robust predictive models that capture the dynamic nature of pricing in the short-term rental market.

The impact of predicting Airbnb prices is significant in the dynamic and ever-growing ecosystem of the sharing economy. Accurate price predictions provide valuable insights and benefits to both hosts and guests, shaping the overall experience and success of the Airbnb platform.

**Empowering hosts:** Accurate price predictions empower hosts to optimize their revenue potential and enhance their overall business performance. They gain valuable insights into expected price trends, market dynamics, and seasonal fluctuations, enabling them to set competitive prices and attract more guests. Precise predictions also help hosts make informed decisions about property upgrades and amenities, further improving the quality of their listings and maximizing profitability.

**Informed decision-making for guests:** Accurate price predictions benefit guests by providing transparency and enabling well-informed decisions. Guests can align their accommodation choices with their preferences and budget, ensuring they get the best value for their money. Transparent pricing information allows guests to plan their trips effectively and make choices that meet their specific needs. This leads to higher guest satisfaction and a better overall experience within the sharing economy.

**Fair and transparent marketplace:** Accurate price predictions contribute to a fair and transparent marketplace within the Airbnb ecosystem. Hosts can confidently set prices that reflect market conditions and demand fluctuations, ensuring fair compensation for their offerings. Guests can trust that the prices they see are aligned with market realities, fostering trust and confidence in the platform. This transparency and fairness create a positive marketplace experience for both hosts and guests.

**Market efficiency and growth:** The prediction of Airbnb prices has broader implications for the sharing economy as a whole. Accurate price predictions optimize pricing strategies, improving resource utilization and reducing waste. Informed guest decision-making leads to a more sustainable use of accommodations and better allocation of resources. By fostering a fair and transparent marketplace, accurate price predictions contribute to the long-term success and growth of the sharing economy.

Accurate prediction of Airbnb prices plays a pivotal role in empowering hosts, facilitating informed guest choices, ensuring a fair marketplace, and advancing the sharing economy. It is a key component in creating a thriving ecosystem that benefits all participants and enhances the overall Airbnb experience.

**Innovation:** Our research hopefully brought meaningful innovation and insight into the prediction and analysis of prices in the short term rental market, and the broader property market as a whole. Our research highlights the importance of certain features that impact demand and price the most in regards to an Airbnb listing. Whether it be the listing's amenities such as the number of rooms or type of rooms available to also how far a listing is to a certain popular location, also we incorporated sentiment analysis to capture the less tangible items of a listing. Our sentiment analysis model was able to analyze reviews and attach a quantitative review score to incorporate how those reviews can impact the price and relative demand of a listing. Furthermore, as the number of Airbnb listings and available data points continue to grow, traditional methods might struggle to handle the sheer volume of information. Machine learning models, such as ours, can scale effortlessly, accommodating large datasets and maintaining predictive accuracy. Businesses that embrace machine learning for predicting Airbnb rental prices gain a competitive edge. By leveraging these advanced techniques, they can fine-tune pricing strategies, enhance customer experiences, and optimize resource allocation, ultimately leading to improved profitability.

## 6.2 FUTURE WORK

In the future, we intend to expand our analysis and refine our predictive capabilities. While our existing methodologies have yielded valuable insights, we encountered limitations that prompted us to adapt our approach.

First, the utilization of the pre-trained RoBERTa (Robustly Optimized BERT approach) model for sentiment analysis was planned. However, due to resource constraints and limited RAM availability, we were unable to incorporate the RoBERTa model into our current framework. Despite this setback, we remain committed to enhancing our sentiment analysis capabilities and leveraging RoBERTa's robustness to achieve more accurate sentiment classification from Airbnb reviews.

We recognize the need to enhance our location features assessment. In future iterations, we aim to incorporate a distance check mechanism to account for listings whose nearest neighbors are located a significant distance away. This enhancement will provide a better understanding of the spatial relationships within the data and contribute to a more thorough analysis of listing locations.

We plan to integrate a neural network regression model for predicting Airbnb prices. Neural networks have demonstrated remarkable proficiency in capturing intricate patterns and relationships in data.

Lastly, our exploration extends to presenting data through mapping. We intend to provide visualizations that express listings per neighborhood, crime rates, and other pertinent information, enhancing our data exploration and aiding guests and hosts in gaining a full overview of the Airbnb market dynamics.

8. APPENDIX

Table 8: Airbnb calendar dataset description.

| Column Name | Data Type | Description |
|---|---|---|
| Listing_id | Integer | ID of Airbnb listing. |
| Date | Date | Calendar date for reservations. |
| Available | Boolean | Boolean value for availability of booking. |
| Price | Integer | FPrice per night of booking or reservation. |
| Adjusted_price | Integer | Adjusted price per night of booking or reservation. |
| Minimum_nights | Integer | Minimum number of nights for booking. |
| Maximum_nights | Integer | Maximum number of nights for booking. |

Table 9: Airbnb review dataset description.

| Column Name | Data Type | Description |
|---|---|---|
| Listing_id | Integer | ID of Airbnb listing. |
| Id | Integer | ID of review. |
| Date | Date | Date of review for listing. |
| Reviewer_id | Integer | ID of reviewer. |
| Reviewer_name | Text | Name of reviewer. |
| Comments | Text | Comments from reviewer. |

Table 10: Airbnb listing dataset description.

| Field | Type | Description |
|---|---|---|
| Id | Integer | Airbnb's unique identifier for the listing. |
| Listing_url | Text | URL of given listing. |
| Scrape_id | Integer | Inside Airbnb ID of "Scrape" this was part of. |
| Last_scraped | Date | The date and time this listing was "scraped". |
| Source | Text | How the listing was found when searching. "Neighborhood search" means that the listing was found by searching the city, while "previous scrape" means that the listing was seen in another scrape performed in the last 65 days. |
| Name | Text | Name of the listing. |
| Description | Text | Detailed description of the listing. |
| Neighborhood_overview | Text | Host's description of the neighborhood. |
| Picture_url | Text | URL to the Airbnb hosted regular sized image for the listing. |
| Host_id | Integer | Airbnb's unique identifier for the host/user. |
| Host_url | Text | The Airbnb page for the host. |
| Host_name | Text | Name of the host. Usually just the first name(s). |
| Host_since | Date | The date the host/user was created. For hosts that are Airbnb guests this could be the date they registered as a guest. |
| Host_location | Text | The host's self reported location. |
| Host_about | Text | Description about the host. |
| Host_response_time | Integer | Typical host response time. One of (a few days or more, within a day, within a few hours, within an hour). |
| Host_response_rate | Integer | Percentage that host responds on time. |
| Host_acceptance_rate | Integer | That rate at which a host accepts booking requests. |
| Host_is_superhost | Boolean | Value determining if the host is considered Superhost or not. |
| Host_thumbnail_url | Text | URL of Airbnb host thumbnail photo (small version). |
| Host_picture_url | Text | URL of Airbnb host thumbnail photo (medium version). |
| Host_neighbourhood | Text | Neighborhood host is located in. |
| Host_listings_count | Text | The number of listings the host has (per Airbnb calculations). |
| Host_total_listings_count | Text | The number of listings the host has (per Airbnb calculations). |
| Host_verifications | Text | Text array of all the ways the host has been verified, |

| | | such as email, phone, government_id, etc. |
|---|---|---|
| Host_has_profile_pic | Boolean | Value determining if the host has a profile picture. |
| Host_identity_verified | Boolean | Value determining if host has verified identity. |
| Neighbourhood | Text | Neighborhood listing is located in. |
| Neighbourhood_cleansed | Text | Approximate zip-code listing is located in. Town for non-U.S. location. |
| Neighbourhood_group_cleansed | Text | Cleansed version of neighborhood group. All currently null. |
| Latitude | Integer | Uses the World Geodetic System (WGS84) projection for latitude and longitude. |
| Longitude | Integer | Uses the World Geodetic System (WGS84) projection for latitude and longitude. |
| Property_type | Text | Property type of listing such as house, guesthouse, boat, etc. |
| Room_type | Text | Room type of listing. One of (Entire home/apt, Private room, Hotel room, Shared room). |
| Accommodates | Integer | Number of maximum guests listing can accommodate. |
| Bathrooms | Integer | The number of bathrooms in the listing. |
| Bathrooms_text | Text | The number of bathrooms in the listing. On the Airbnb web-site, the bathrooms field has evolved from a number to a textual description. For older scrapes, bathrooms are used. |
| Bedrooms | Integer | The number of bedrooms. |
| Beds | Integer | The number of bed(s). |
| Amenities | Text | Text array of all the amenities available at listing such as TV, wifi, air conditioning, etc. |
| Price | Integer | Price per night of booking or reservation. |
| Minimum_nights | Integer | Minimum number of night stay for the listing. |
| Maximum_nights | Integer | Maximum number of night stay for the listing. |
| Minimum_minimum_nights | Integer | The smallest minimum_night value from the calendar (looking 365 nights in the future). |
| Maximum_minimum_nights | Integer | The largest minimum_night value from the calendar (looking 365 nights in the future). |
| Minimum_maximum_nights | Integer | The smallest maximum_night value from the calendar (looking 365 nights in the future). |
| Maximum_maximum_nights | Integer | The largest maximum_night value from the calendar (looking 365 nights in the future). |
| Minimum_nights_avg_ntm | Numeric | The average minimum_night value from the calendar (looking 365 nights in the future). |
| Maximum_nights_avg_ntm | Numeric | The average maximum_night value from the calendar (looking 365 nights in the future). |
| Calendar_updated | Date | How long ago the calendar for listing was last updated. |
| Has_availability | Boolean | Value determining if listing has availability. |
| Availability_30 | Integer | Number of nights available in the next (last) 30 days. |
| Availability_60 | Integer | Number of nights available in next (last) 60 days. |
| Availability_90 | Integer | Number of nights available in the next (last) 90 days. |
| Availability_365 | Integer | Number of nights available in the next (last) 365 days. |
| Calendar_last_scraped | Date | Date of latest calendar web scraping from Airbnb website. |
| Number_of_reviews | Integer | The number of reviews the listing has. |
| Number_of_reviews_ltm | Integer | The number of reviews the listing has (in the last 12 months). |
| Number_of_reviews_l30d | Integer | The number of reviews the listing has (in the last 30 days). |
| First_review | Date | The date of the first/oldest review. |
| Last_review | Date | The date of the last/newest review. |
| Review_scores_rating | Integer | Score on how guests felt overall during their stay. |
| Review_scores_accuracy | Integer | Score on how guests felt that the listing page represented the space. |
| Review_scores_cleanliness | Integer | Score on how guests felt that the space was clean and tidy during their stay. |
| Review_scores_checkin | Integer | Score on how guests felt that check-in went through smoothly. |
| Review_scores_communication | Integer | Score on how guests felt that the host communicated before and during their stay. |
| Review_scores_location | Integer | Score on how guests felt about the location/neighborhood of the listing and how close it was to other desired locations or attractions. |
| Review_scores_value | Integer | Score on how guests felt about the overall value of the listing for the price. |
| License | Text | Rental license number for host. |
| Instant_bookable | Boolean | Value determining if instant booking is available. |
| Calculated_host_listings_count | Integer | The number of listings the host has in the current scrape, in the city/region geography. |
| Calculated_host_listings_count_entire_homes | Integer | The number of Entire home/apt listings the host has in the current scrape, in the city/region geography. |
| Calculated_host_listings_count_private_rooms | Integer | The number of Private room listings the host has in the current scrape, in the city/region geography. |
| Calculated_host_listings_count_shared_rooms | Integer | The number of Shared room listings the host has in the current scrape, in the city/region geography. |
| Reviews_per_month | Integer | The number of reviews the listing has over the lifetime |

|  |  | of the listing. |